%% file: main.tex
\documentclass[10pt,twocolumn,letterpaper]{article}

\usepackage{iccv}              

\input{preamble}

\definecolor{iccvblue}{rgb}{0.21,0.49,0.74}
\usepackage[pagebackref,breaklinks,colorlinks,allcolors=iccvblue]{hyperref}
\usepackage[symbol]{footmisc}

\def\paperID{4855} 
\def\confName{ICCV}
\def\confYear{2025}

\begin{document}

\title{Stereo Any Video: Temporally Consistent Stereo Matching}


\author{
  Junpeng Jing \quad
  Weixun Luo \quad
  Ye Mao\footnotemark[2]  \quad
  Krystian Mikolajczyk \\
  Imperial College London \\
  {\tt\small \url{https://tomtomtommi.github.io/StereoAnyVideo/}}
}

\twocolumn[{
\renewcommand\twocolumn[1][]{#1}
\maketitle
\begin{center}
    \centering
    \vspace{-.6em}
    \includegraphics[width=0.99\linewidth,page=1]{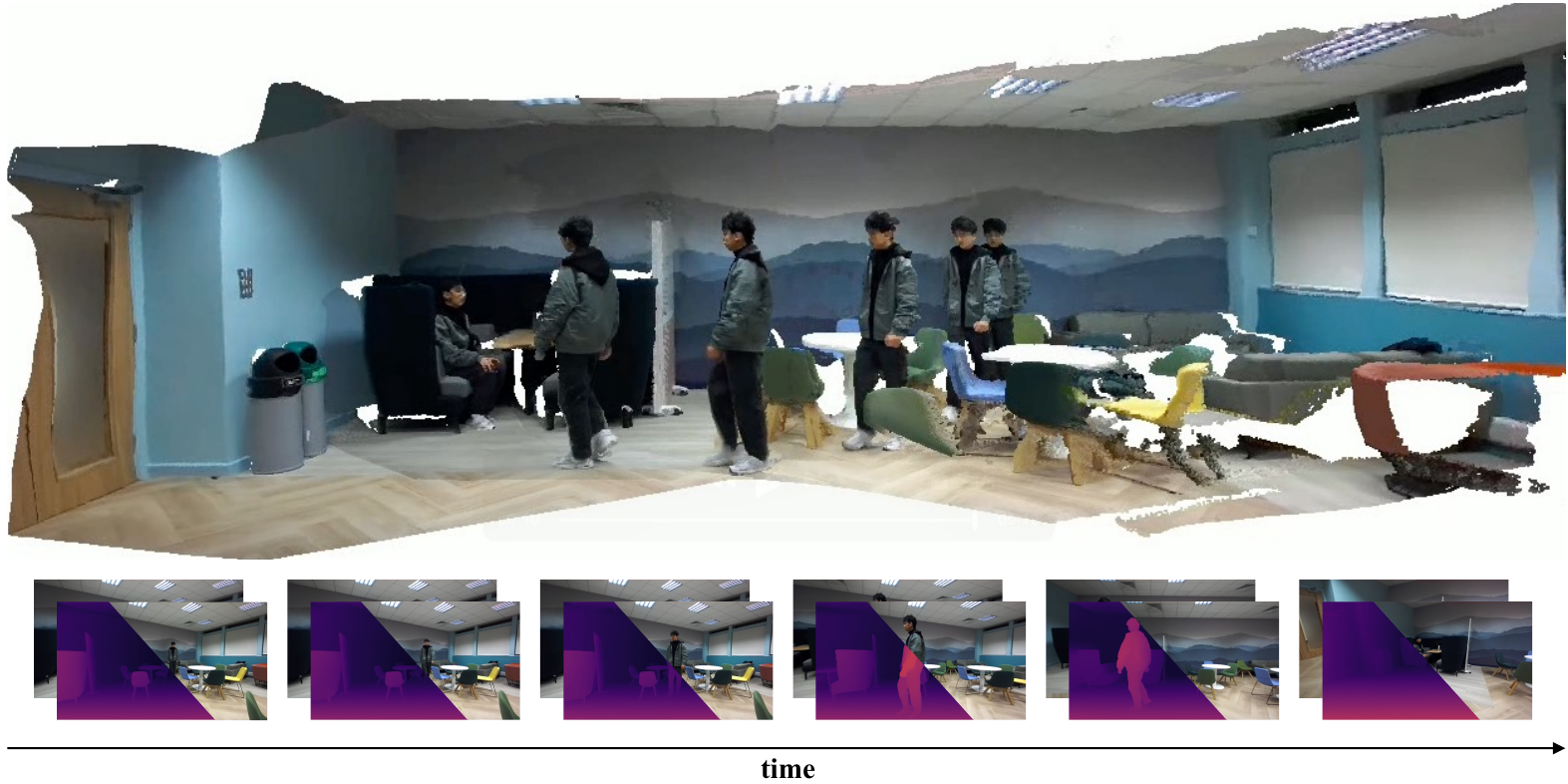}
    \vspace{-0.8em}
    \captionof{figure}{Prediction on a dynamic real-world stereo video. Our method produces temporally consistent and accurate disparities without flickering, enabling the reconstruction of a dense and clean metric point cloud for dynamic scene. See video for better visualization.}
    \label{fig:figure1}
\end{center}
}]

\footnotetext[2]{Corresponding author: \texttt{ye.mao21@imperial.ac.uk}}

\begin{abstract}
This paper introduces \textbf{Stereo Any Video}, a powerful framework for video stereo matching. It can estimate spatially accurate and temporally consistent disparities without relying on auxiliary information such as camera poses or optical flow. The strong capability is driven by rich priors from monocular video depth models, which are integrated with convolutional features to produce stable representations. To further enhance performance, key architectural innovations are introduced: all-to-all-pairs correlation, which constructs smooth and robust matching cost volumes, and temporal convex upsampling, which improves temporal coherence. These components collectively enhance robustness, accuracy, and temporal consistency, establishing a new standard in video stereo matching. Extensive experiments demonstrate that our method achieves state-of-the-art performance across multiple datasets both qualitatively and quantitatively in zero-shot settings, as well as strong generalization to real-world indoor and outdoor scenarios.
\end{abstract}

\section{Introduction}
Video stereo matching is a task to reconstruct a 3D scene from a stereo camera. It estimates disparity by identifying the horizontal correspondences between rectified left and right images at each time step. Consistent and accurate disparity estimation is crucial for reconstructing and understanding real-world 3D scenes, benefiting various downstream tasks such as autonomous driving, robotic navigation, and virtual/augmented reality. 

With the advances in deep learning, stereo matching has seen significant progress \cite{lipson2021raft, li2022practical, xu2023iterative, wang2024selective}, achieving promising performance on public benchmarks \cite{middlebury, eth3d, kitti}. However, most existing approaches are designed for image-based estimation and do not explicitly leverage temporal information in video sequences. When applied directly to video data, these methods often suffer from temporal inconsistencies, leading to noticeable flickering and artifacts in disparity maps and reconstructed point clouds. To mitigate this, recent video-based methods incorporate additional temporal cues, such as camera poses or optical flow, through predefined models or learnable mechanisms. For example, CODD \cite{li2023temporally} employs a pose network to estimate relative camera movements and align adjacent disparities within a fusion framework. Similarly, TemporalStereo \cite{zhang2023temporalstereo} and TC-Stereo \cite{zeng2024temporally} follow this paradigm with variations in architectural design and alignment operations. Another approach, BiDAStereo \cite{jing2024matchstereovideos}, integrates optical flow \cite{teed2020raft} for temporal feature alignment and introduces a temporal aggregation module to propagate information across time.

Despite these advances, existing video methods remain vulnerable to dynamic scenes, camera movement, and the accuracy of auxiliary information. In real-world scenarios, dynamic objects undergo complex motions and deformations that violate traditional multi-view constraints \cite{hartley2003multiple}, disrupting translational invariance in disparity estimation. Moreover, models relying on auxiliary cues are constrained by the performance of camera pose or optical flow estimation modules. In environments with complex camera movements, these estimation modules degrade, becoming bottlenecks that limit the robustness of video stereo methods.

To address these limitations, we propose Stereo Any Video, a robust framework for video-based stereo matching without relying on camera pose or optical flow. Inspired by recent findings in video generation \cite{luo2024Enhance-A-Video, si2025repvideorethinkingcrosslayerrepresentation}, which emphasize the role of robust feature representations in ensuring temporal coherence, we take a new perspective to design the framework for consistency: feature robustness and stability. Specifically, during feature extraction, we integrate convolutional cues with a frozen monocular video depth foundation model \cite{chen2025videodepthanythingconsistent} trained on million-scale data. This ensures robust feature maps with rich priors. For correspondence matching, we introduce an all-to-all-pair correlation, which enforces smooth and reliable matching by jointly considering bidirectional correspondences from both target and reference views. For cost aggregation, we propose temporal convex upsampling within a fully temporal-based recurrent unit, playing a key role in maintaining temporal consistency. As shown in \Cref{fig:figure1}, our method demonstrates a strong ability to produce accurate and consistent disparities, benefiting the reconstruction for dynamic scenes.

The contributions of our work can be summarized as follows: (1) Stereo Any Video, a framework that integrates monocular video depth priors to achieve accurate and temporally consistent video stereo matching. (2) A novel all-to-all-pair correlation module and a temporal convex upsampling mechanism, ensuring robust and stable cost volume representations and aggregation. (3) State-of-the-art performance on diverse benchmarks, setting a new standard for video stereo matching.

\begin{figure*}[t]
   \begin{center}
   \includegraphics[width=.9\linewidth]{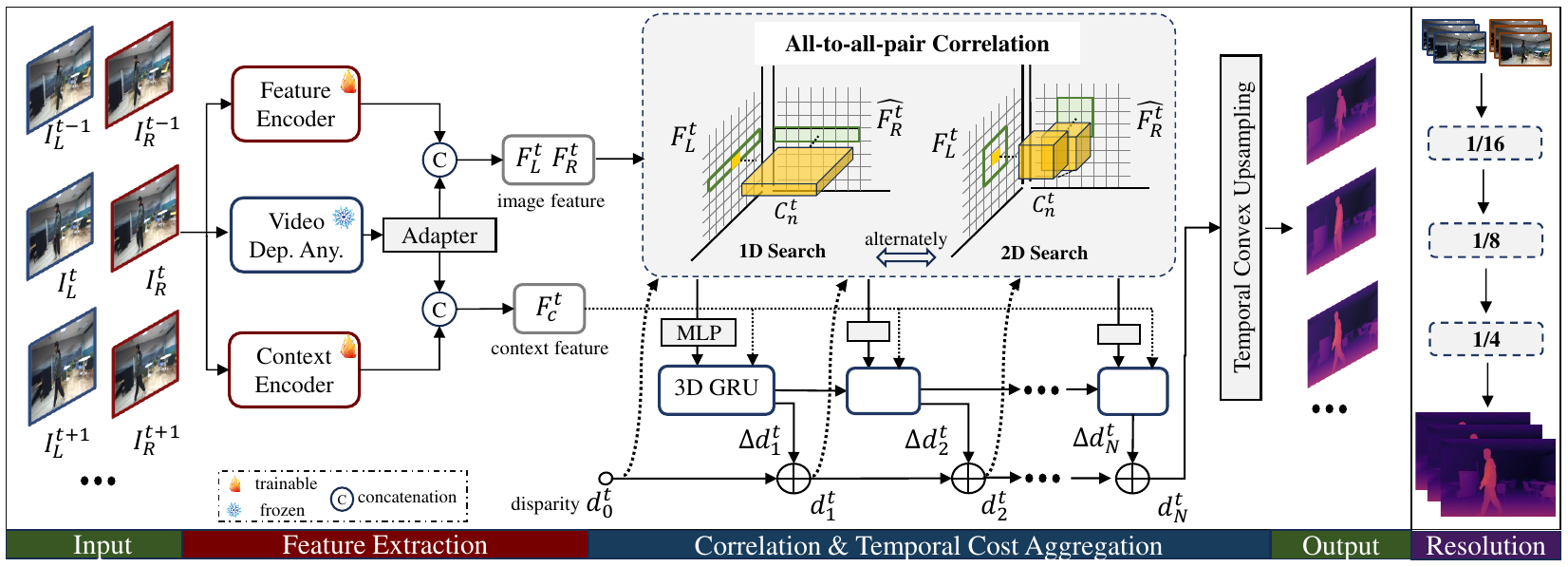}
   \end{center}
   \vspace{-1em}
    \caption{Overview of Stereo Any Video. Given a stereo video sequence as input, the model first extracts features and context information using trainable convolutional encoders and a frozen monocular video depth encoder (Video Depth Anything \cite{chen2025videodepthanythingconsistent}). At each iteration, an all-to-all-pair correlation computes feature correlations, followed by a multilayer perception (MLP) encoder for compact representation. Disparities are iteratively refined using a 3D Gated Recurrent Unit (GRU), integrated with cost volume features, and upsampled via a temporal convex upsampling layer. This process is repeated within a cascaded pipeline to progressively recover full-resolution disparities.} 
    \vspace{-1em}
   \label{fig:framework}
\end{figure*}

\section{Related Work}
In this section, we provide a brief literature review of stereo matching methods in the deep learning era. We first discuss image-based stereo matching methods, followed by video-based methods, and finally, we review recent advancements leveraging foundation priors.

\noindent \textbf{Image-based Stereo Matching.} Since the introduction of the first end-to-end stereo network \cite{mayer2016large}, deep learning-based stereo matching has seen substantial advancements. Early methods typically construct 3D \cite{pang2017cascade, liang2018learning, guo2019group, xu2020aanet, tankovich2021hitnet} or 4D \cite{kendall2017end, chang2018pyramid, zhang2019ga, yang2019hierarchical, cheng2020hierarchical, gu2020cascade,xu2023accurate} cost volumes from left and right image features, followed by 2D or 3D convolutions for cost aggregation. Inspired by RAFT \cite{teed2020raft}, iterative refinement has become the mainstream approach. Representative advancements include multilevel recurrent units \cite{lipson2021raft}, adaptive correlation \cite{li2022practical}, geometry-encoding volumes \cite{xu2023iterative, xu2025igev++}, and frequency information selection \cite{wang2024selective}, all of which have significantly improved benchmark performance \cite{middlebury, eth3d, kitti}. Transformer-based methods \cite{li2021revisiting, guo2022context, su2022chitransformer, weinzaepfel2023croco, xu2023unifying} have further refined stereo matching by modeling token correspondence and capturing long-range dependencies. Beyond accuracy, several methods focus on real-world challenges such as efficiency \cite{wang2019anytime, xu2021bilateral, xu2023cgi, guo2024lightstereochannelboostneed}, domain generalization \cite{pang2018zoom, chuah2022itsa, zhang2022revisiting, chang2023domain, rao2023masked}, and robustness \cite{shen2021cfnet, ucfnet, Jing_2023_ICCV}. However, as all these methods are designed for static image pairs, they lack temporal awareness, leading to inconsistent disparity predictions in video sequences.

\noindent \textbf{Video-based Stereo Matching.} Video stereo matching especially in dynamic scenes presents unique challenges. Early work, such as \cite{zhong2018open}, employed LSTMs with unsupervised learning for stereo videos but did not explicitly incorporate temporal information. CODD \cite{li2023temporally} introduced a per-frame stereo network, a motion network, and a fusion network designed for dynamic scenes. Neighboring-frame disparities are aligned and fused using estimated camera poses. TemporalStereo \cite{zhang2023temporalstereo} extended this approach with a coarse-to-fine network that leverages past context to enhance predictions in challenging scenarios. Cheng \etal \cite{cheng2024stereo} developed a framework tailored for XR systems, reducing computational costs through temporal cost aggregation. TC-Stereo \cite{zeng2024temporally} further introduced a temporal disparity completion mechanism and a temporal state fusion module to exploit temporal information. These methods rely on auxiliary camera motion or pre-defined scene geometry to incorporate temporal information. DynamicStereo \cite{karaev2023dynamicstereo} introduced a transformer-based architecture that integrates temporal, stereo, and spatial attention. BiDAStereo \cite{jing2024matchstereovideos} proposed an optical-flow-based alignment mechanism to propagate temporal information across the entire sequence. Building on this alignment, BiDAStabilizer \cite{jing2024matchstereovideosbidirectional} introduced a lightweight plug-and-play stabilizer network to improve the temporal consistency of existing image-based stereo matching methods.

\noindent \textbf{Stereo Matching with Foundation Priors.} Vision foundation models, such as CLIP \cite{radford2021learning}, the DINO series \cite{caron2021emerging, oquab2023dinov2}, and SAM \cite{kirillov2023segment}, have been widely applied across various vision tasks due to their rich priors. In stereo matching area, FormerStereo \cite{zhang2024learning} proposed feature transformation and C4 space to adapt ViT-based foundation model to stereo pipeline, learning domain invariant representations. AIOStereo \cite{zhou2024all} introduced a selective knowledge transfer module to distill knowledge from multiple heterogeneous vision foundation models into a single stereo matching model. More recent approaches primarily focused on leveraging monocular depth priors from Depth Anything \cite{yang2024depth, yang2024depth2} to enhance stereo performance. For instance, Stereo Anywhere \cite{bartolomei2024stereo} directly incorporated relative depth maps into the stereo network, demonstrating robustness in challenging conditions such as non-Lambertian surfaces. DEFOM-Stereo \cite{jiang2025defomstereodepthfoundationmodel} employed these depth maps alongside a novel scale updater within a recurrent module for improved alignment. FoundationStereo \cite{wen2025foundationstereozeroshotstereomatching} extracted feature maps from Depth Anything and introduced a side-tuning adapter to adapt monocular features for stereo backbone. MonSter \cite{cheng2025monstermarrymonodepthstereo} utilized both depth maps and feature maps, and proposed a dual-branched refinement module to fuse stereo and monocular cues effectively. These methods highlight the effectiveness of foundation priors in image-based stereo tasks. In our work, we build upon this direction by investigating video-based stereo matching and exploring how foundation priors can enhance consistency in dynamic scenes.

\section{Method}
In this section, we introduce Stereo Any Video, a feed-forward video-based framework. Given a rectified stereo sequence $\{\mathbf{I}^{t}_{L}, \mathbf{I}^{t}_{R}\}_{t\in (1,T)} \in \mathbb{R}^{H\times W\times 3} $ where $T$ denotes the total number of frames, $H$ and $W$ are the frame height and width, respectively, our goal is to estimate a temporally consistent sequence of disparity maps $\{\mathbf{d}^{t}\}_{t\in (1,T)} \in \mathbb{R}^{H\times W} $ aligned with the left view. As shown in \Cref{fig:framework}, our method consists of three key components: feature extraction (\Cref{sec:Feature Extraction}), correlation (\Cref{sec:Correlation}), and temporal cost aggregation (\Cref{sec:Aggregation}).

\subsection{Feature Extraction with Foundation Priors} \label{sec:Feature Extraction}
Unlike image-based tasks, maintaining temporal consistency is the primary challenge in video stereo matching. Recent studies in video generation \cite{luo2024Enhance-A-Video, si2025repvideorethinkingcrosslayerrepresentation} indicate that substantial variations across different frames can lead to unstable feature representations, thereby degrading temporal coherence. Therefore, ensuring feature stability across adjacent frames is crucial for achieving temporally consistent disparity estimation. Towards this, we leverage Video Depth Anything (VDA) \cite{chen2025videodepthanythingconsistent} for feature extraction, which offers two key advantages: (1) trained on internet-scale data, VDA provides rich priors that can enhance disparity estimation in regions with ambiguous stereo correspondence; (2) unlike image-based monocular depth models \cite{yang2024depth, yang2024depth2}, VDA generates temporally stable features through an explicit temporal alignment process before the final prediction head, thereby preserving coherence across frames.

\begin{figure}[t]
   \begin{center}
   \includegraphics[width=.95\linewidth]{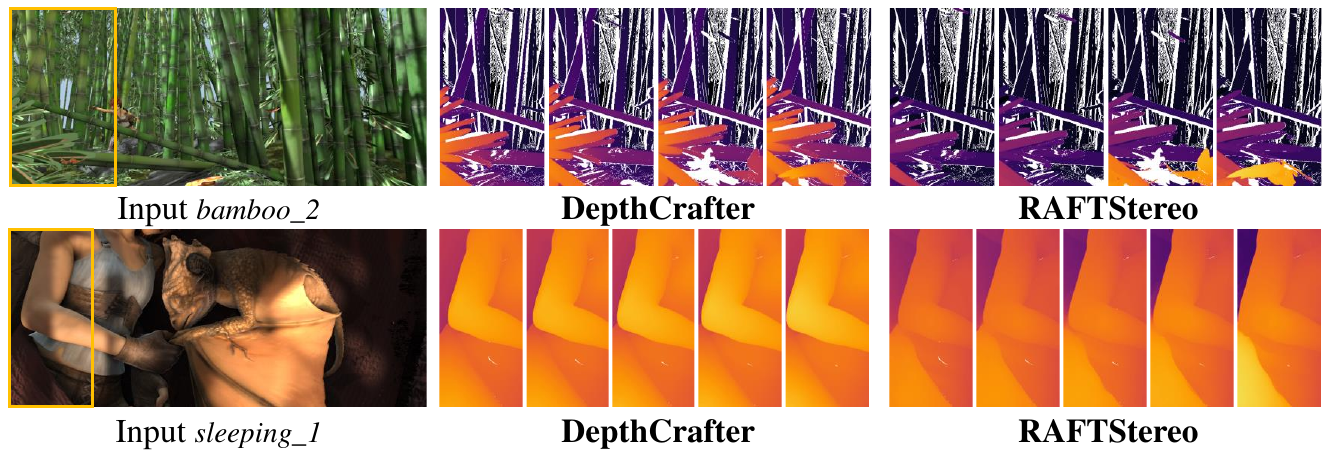}
   \end{center}
   \vspace{-1.5em}
    \caption{Qualitative comparison of DepthCrafter \cite{hu2024-DepthCrafter} and RAFTStereo \cite{lipson2021raft} on Sintel dataset \cite{sintel}.} 
    \vspace{-1.4em}
    \label{fig:motivation}
\end{figure}

Besides, we investigate whether the depth maps from monocular video depth models can further improve stereo performance. We compare DepthCrafter \cite{hu2024-DepthCrafter}, a representative monocular video model, with RAFTStereo \cite{lipson2021raft}, an image-based stereo model. Using open-source weights, we evaluate their performance on Sintel \cite{sintel}, as shown in \Cref{tab:motivation} and \Cref{fig:motivation}. Specifically, we first obtain relative depth maps from DepthCrafter (without scale and shift) and compute disparities using RAFTStereo on the same sequence. The relative depth maps are then aligned to RAFTStereo’s disparities, and objective metrics are computed against the ground truth. Additionally, we conduct a subjective user study (see supplementary material for details). As observed, while DepthCrafter produces visually plausible results—reflected in better human scores—its spatial and temporal errors are significantly larger than those of RAFTStereo. This suggests that its relative depth is ``consistent but wrong", making it unreliable for directly improving consistency within a stereo framework. Therefore, we only utilize its features as valuable priors. 

As shown in \Cref{fig:framework}, our feature extraction module consists of two trainable encoders, each employing a series of residual blocks \cite{lipson2021raft} to extract convolutional image features and context features. To integrate VDA priors, we adopt a shallow convolutional adapter \cite{wen2025foundationstereozeroshotstereomatching} that downsamples depth features from the frozen VDA. The extracted depth features are then concatenated with convolutional features to form $\left \{\mathbf{F}_{L} ,\mathbf{F}_{R} \right \}$ and $\mathbf{F}_{C}$, providing more stable feature representations for subsequent processing.

\begin{table}[t]
\centering
\setlength{\tabcolsep}{6.pt}
\footnotesize
\caption{Quantitative comparison of DepthCrafter \cite{hu2024-DepthCrafter} and RAFTStereo \cite{lipson2021raft} on Sintel dataset \cite{sintel}.}
\vspace{-1em}
\begin{tabular}{l|ccc}
\toprule
\multirow{2}{*}{\qquad Method} & \multicolumn{3}{c}{Sintel Clean}    \\
\cmidrule(lr){2-4}
 & EPE $\downarrow $ & TEPE $\downarrow $ & MOS (human) $\downarrow $ \\
\midrule
DepthCrafter \cite{hu2024-DepthCrafter}  & 8.68 & 2.08 & 0.212  \\
RAFTStereo \cite{lipson2021raft}     & 1.42 & 0.84 & 0.933 \\
\bottomrule
\end{tabular}
\label{tab:motivation}
\vspace{-1.1em}
\end{table}

\subsection{All-to-all-pair Correlation}
\label{sec:Correlation}
At the $n$-th iteration for each time step, the right feature $\mathbf{F}_{R}$ is first warped toward the left feature $\mathbf{F}_{L}$ using the previous disparity ${d}_{n-1}$, producing $\mathbf{\hat{F}}_{R}$. Previous methods \cite{lipson2021raft, xu2023iterative} typically construct cost volumes $\mathbf{C}_n$ via all-pair correlation, employing a lookup mechanism to search for potential matches within a local radius. In \cite{li2022practical}, this approach was extended to an alternating local 1D and 2D search:
\begin{equation}
    \mathbf{C}_n(x, y) = \Big\langle \mathbf{F}_L(x, y), \hat{\mathbf{F}}_R(x+r_x, y+r_y) \Big\rangle,
\end{equation}
where $(x,y)$ denotes the position in $\mathbf{F}_{L}$, $\langle \cdot, \cdot \rangle$ represents the channel-wise product operation, and $(r_x, r_y)$ defines the search range, typically set as $(r_x, r_y) \in \{ (\pm 4,0), (\pm1,\pm1)\}$. The fundamental principle of these methods is to construct one-directional correspondence, where a reference point is selected and directly matched with target points via multiplication. Inspired by recent advances in point tracking \cite{cho2024local}, we introduce bidirectional correspondence and develop a local all-to-all-pair correlation. Instead of computing correlation in a fixed direction, we calculate the similarity between every pair of potential matching points within two search windows, formulated as:
\begin{equation}
    \mathbf{C}_n(x, y) = \Big\langle \mathbf{F}_L(x+r_x, y+r_y), \hat{\mathbf{F}}_R(x+r_x, y+r_y) \Big\rangle.
    \label{eq:1}
\end{equation}

This approach enhances match verification and reduces ambiguity \cite{lowe2004distinctive, rocco2018neighbourhood}. While BiDAStereo \cite{jing2024matchstereovideos} constructs temporal cost volumes by aligning adjacent frames for consistency, its performance is significantly constrained by the limitations of optical flow. In contrast, our method emphasizes robust spatial correlation, leveraging dense correspondences to enforce matching smoothness and enhance consistency and robustness across neighboring points \cite{truong2020gocor, truong2020glu}.

\begin{figure}[t]
   \begin{center}
   \includegraphics[width=.95\linewidth]{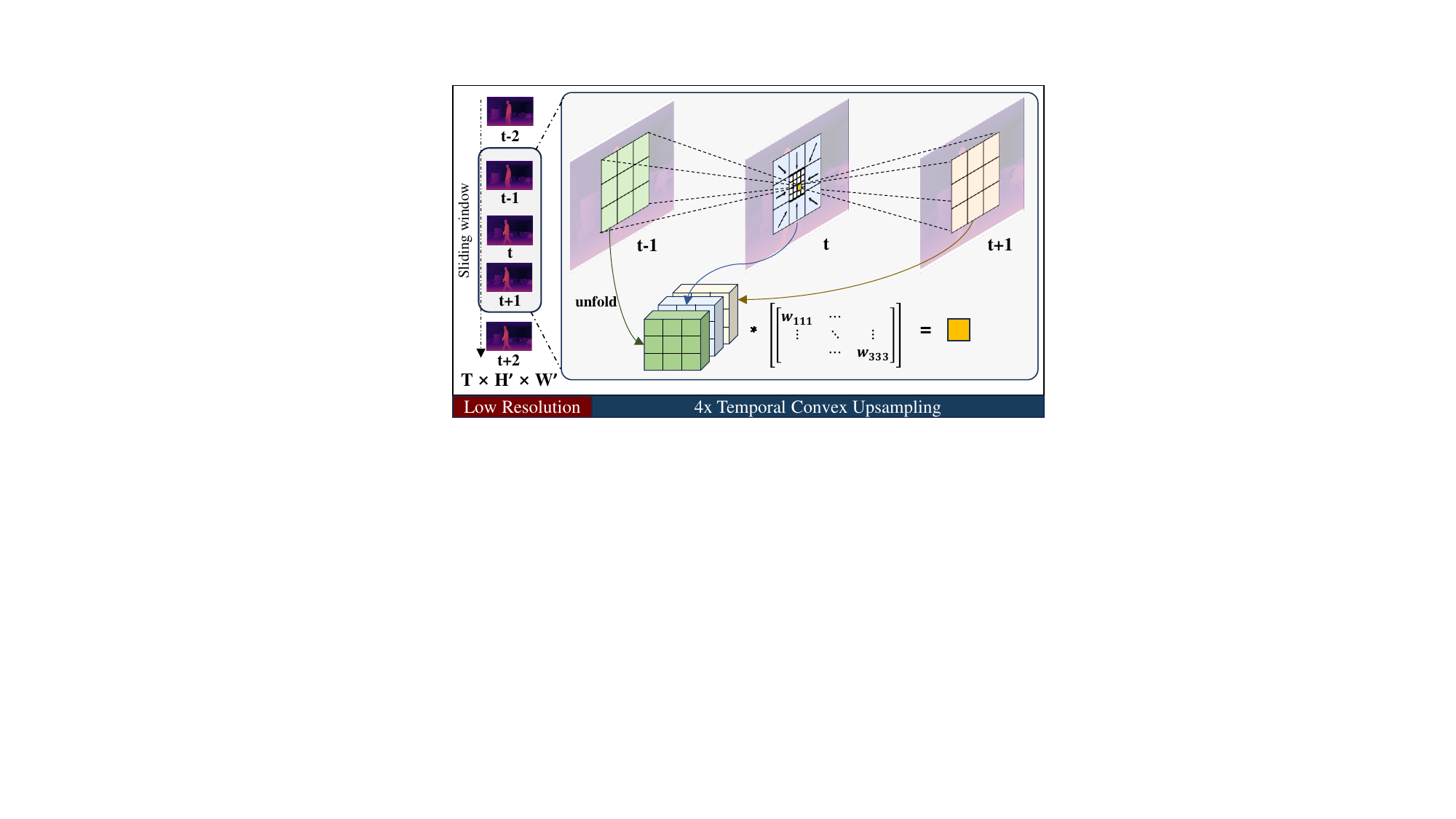}
   \end{center}
   \vspace{-1.6em}
    \caption{Workflow of the temporal convex upsampling. Each pixel of the high resolution disparity (yellow box) is calculated via its 27 low resolution neighbors using learnable weights.} 
    \vspace{-1.2em}
    \label{fig:convex}
\end{figure}

\begin{table*}[tb] 
\caption{Zero-shot generalization results. All methods were trained only on SceneFlow \cite{mayer2016large}. The representative image-based and video-based stereo matching models are compared on Sintel \cite{sintel}, Dynamic Replica \cite{karaev2023dynamicstereo}, Infinigen SV \cite{jing2024matchstereovideosbidirectional}, and Virtual KITTI2 \cite{cabon2020vkitti2}. The weights and parameters are fixed for evaluation. The \textbf{best} and \underline{second-best} results are highlighted.}
\vspace{-.8em}
\centering
\setlength{\tabcolsep}{6.pt}
\footnotesize
\begin{tabular}{l|ccc|ccc|ccc|ccc}
\toprule
\multirow{2}{*}{\qquad \;  Method} & \multicolumn{3}{c}{Sintel Final}  & \multicolumn{3}{|c}{Dynamic Replica}   & \multicolumn{3}{|c}{Infinigen SV} & \multicolumn{3}{|c}{Virtual KITTI2}   \\
\cmidrule(lr){2-13}
 & TEPE$\downarrow $ & $\delta_{1px}^{t}\downarrow $ & EPE$\downarrow $  & TEPE$\downarrow $ & $\delta_{1px}^{t}\downarrow $ & EPE$\downarrow $  & TEPE$\downarrow $ & $\delta_{1px}^{t}\downarrow $ & EPE$\downarrow $   & TEPE$\downarrow $ & $\delta_{1px}^{t}\downarrow $ & EPE$\downarrow $ \\
\midrule
\rowcolor{gray!20} \multicolumn{13}{l}{\hspace{0pt}\textit{Image-based}}\\
RAFTStereo \cite{lipson2021raft}   & 2.09 & 13.70 & 4.89 & 0.145 & 2.02 & 0.38 & 2.33 & 19.80 & {3.77} & 1.25 & 18.39 & 1.45 \\
IGEVStereo \cite{xu2023iterative}  & 1.74 & 12.65 & 2.44 & 0.081 & 0.80 & \underline{0.20} & 2.35 & 19.33 & 3.81 & 1.42 & 18.88 & 1.39 \\
Selective-IGEV \cite{wang2024selective} & 1.96 & 12.95 & 2.82 & 0.133 & 1.29 & 0.29 & 2.85 & 21.83 & 5.69 & 1.41 & 19.10 & {1.38} \\
DEFOM-Stereo \cite{jiang2025defomstereodepthfoundationmodel} & 2.33 & 12.43 & 2.38 & \underline{0.080} & 0.75 & \textbf{0.13} & 2.01 & 18.91 & 3.24 & 1.07 & 16.63 & 1.34  \\
MonSter \cite{cheng2025monstermarrymonodepthstereo} & 1.70 & \underline{11.06} & 3.30 & 0.157 & 0.86 & 0.45 & \textbf{1.65} & 15.73 & \textbf{2.27} & \textbf{0.73} & \textbf{10.93} & \textbf{0.67}  \\
\rowcolor{gray!20} \multicolumn{13}{l}{\hspace{0pt}\textit{Video-based}}\\
DynamicStereo \cite{karaev2023dynamicstereo}  & 1.43 & 11.80 & 4.26 & 0.083 & \underline{0.73} & {0.24} & \underline{1.98} & \underline{15.68} & 4.67 & 1.06 & 15.63 & 1.47  \\
BiDAStereo \cite{jing2024matchstereovideos}   & \underline{1.26} & {11.65} & \underline{1.99} & 0.091 & 1.04 & 0.65 & 1.99 & 16.50 & 6.52 & {1.02} & {14.82} & 1.43  \\
Stereo Any Video (ours) & \textbf{1.07} & \textbf{9.98} & \textbf{1.75} & \textbf{0.057} & \textbf{0.67} & 0.25 & \textbf{1.65} & \textbf{14.27} & \underline{3.16} & \underline{0.74} & \underline{12.45} & \underline{1.01}  \\
\bottomrule
\end{tabular}
\vspace{-1.2em}
\label{tab:zeroshot}
\end{table*}

\subsection{Temporal Cost Aggregation} \label{sec:Aggregation}
At the $n$-th iteration, the all-to-all-pair correlation expands the potential matching points in $\mathbf{C}_n$, increasing its dimensionality from $\mathbb{R}^{H'\times W'\times (2r_x+1)(2r_y+1)}$ to $\mathbb{R}^{H'\times W'\times (2r_x + 1)^2(2r_y + 1)^2}$, where $H'$ and $W'$ denote the height and width of frame feature, respectively. We first encode $\mathbf{C}_n$ into a compact representation $\mathbf{E}_n \in \mathbb{R}^{H'\times W'\times L}$ using a MLP:
\begin{equation}
\mathbf{E}_n = \text{MLP}(\mathbf{C}_n).
\end{equation}
Next, we employ a 3D-GRU \cite{karaev2023dynamicstereo} to iteratively refine the disparity for each frame while preserving temporal consistency. The GRU aggregates information using separable 3D convolutions across both spatial and temporal dimensions. At iteration $n$, it takes as input the encoded cost volume $\mathbf{E}_n$, context feature $\mathbf{F}_c$, and the disparities from the previous iteration $d_{n-1}$. To enhance representation, we integrate super kernels \cite{jing2024matchstereovideos}, temporal and spatial attention \cite{karaev2023dynamicstereo} in this process. The disparity update in 3D-GRU is as follows:
\begin{align}
    & x_n = \big[ \text{Conv2d}(\mathbf{E}_n), \text{Conv2d}(d_{n-1}), d_{n-1}, \mathbf{F}_c \big], \label{eq:2} \\
    & \widetilde{x_n} = \text{Attention}(x_n), \\
    & z_n = \sigma \Big( \text{Conv3d}_z \big( [h_{n-1}, \widetilde{x_n}] \big) \Big), \label{eq:3} \\
    & r_n = \sigma \Big( \text{Conv3d}_r \big( [h_{n-1}, \widetilde{x_n}] \big) \Big), \label{eq:4} \\
    & q_n = \tanh \Big( \text{Conv3d}_q \big( [h_{n-1} \odot r_n, \widetilde{x_n}] \big) \Big), \label{eq:5} \\
    & h_n = (1 - z_n) \odot h_{n-1} + z_n \odot q_n, \label{eq:7} \\
    & d_n = d_{n-1} + \text{Conv3d}(h_n), \label{eq:8}
\end{align}
where $[ \cdot ]$ is concatenation, $\odot$ denotes element-wise multiplication, and $\sigma$ is the sigmoid activation function.

While prior methods emphasize temporal consistency in correlation and cost aggregation, the upsampling process remains overlooked, despite its crucial role in achieving accurate and temporally consistent disparities. Towards this, we adapt the commonly used convex upsampling approach \cite{teed2020raft} to a temporal variant. In \Cref{fig:convex}, each high-resolution disparity pixel (yellow box) is computed as a convex combination of its $27$ low resolution neighbors across the $3$ frames (green, blue, and faint yellow boxes). The network predicts the corresponding $3\times3\times3$ weights using 3D convolutions, enabling a temporally consistent upsampling:
\begin{align}
    & \mathbf{w} = \text{softmax} \Big(\text{Conv3d}(h_n) \Big),\\
    & D_n^t = \alpha \cdot \sum_{ijk} \mathbf{w}_{ijk} \odot \text{unfold}([d_n^{t-1}, d_n^t, d_n^{t+1}]),
\end{align}
where $\alpha$ is upsampling rate, ${D}_{n}^{t}$ is the upsampled disparity.

\subsection{Loss Function}
Recent works have introduced OPW loss \cite{NVDS} and TGM loss \cite{chen2025videodepthanythingconsistent} in monocular video depth estimation to enhance temporal consistency. However, we found out that these losses impose a trade-off between accuracy and consistency without yielding substantial improvement in the video stereo task. Thus, we opt for a purely image-based $L_1$ loss while still achieving strong temporal consistency as:
\begin{equation}
\mathcal{L} = \sum_{t=1}^{T} \sum_{n=1}^{N} 
\gamma^{N - n} || {D}_{\mathrm{gt}}^{t} - {D}_{n}^{t} ||,
\end{equation}
where $T$ is the number of frames, $N$ is the number of iterations, and $\gamma$ is a weighting coefficient set to 0.9. Here, ${D}_{\mathrm{gt}}^{t}$  denotes the ground truth disparity for the $t$-th frame. The final output is obtained from the last iteration prediction.

\section{Experiments}
In this section, we first introduce the datasets and metrics used in the evaluation, as well as implementation details. We then present the zero-shot generalization capabilities of our method, followed by ablation studies to validate the efficacy of each component in our framework. Please refer to the supplementary material for more details.


\begin{table*}[tb] 
\caption{Quantitative comparisons on Spring \cite{Mehl2023_Spring}, Sintel \cite{sintel}, and KITTI Depth \cite{kittidepth} where methods are allowed to train on any existing datasets. Model weights remain fixed during evaluation.  Training datasets sharing the same domain as the test sets are highlighted. ``--" represents the test set has been used in training. Abbreviations: SF – SceneFlow \cite{mayer2016large}, DR – Dynamic Replica \cite{karaev2023dynamicstereo}, ISV – Infinigen SV \cite{jing2024matchstereovideosbidirectional}, M – Middlebury \cite{middlebury}, KD – KITTI Depth \cite{kittidepth}, VK – Virtual KITTI2 \cite{cabon2020vkitti2}. The \textbf{best} and \underline{second-best} results are highlighted.}
\vspace{-.6em}
\centering
\setlength{\tabcolsep}{6.pt}
\footnotesize
\begin{tabular}{l|c|c|ccc|ccc|ccc}
\toprule
\multirow{2}{*}{\qquad \;  Method}  & \multirow{2}{*}{Training Data} & \multirow{2}{*}{Scale} & \multicolumn{3}{c}{Spring} & \multicolumn{3}{|c}{Sintel Final}   & \multicolumn{3}{|c}{KITTI Depth}   \\
\cmidrule(lr){4-12}
 &  & & TEPE$\downarrow $ & $\delta_{1px}^{t}\downarrow $ & EPE$\downarrow $ 
 & TEPE$\downarrow $ & $\delta_{1px}^{t}\downarrow $ & EPE$\downarrow $
 & TEPE$\downarrow $ & $\delta_{1px}^{t}\downarrow $ & EPE$\downarrow $ \\
\midrule
\rowcolor{gray!20} \multicolumn{12}{l}{\hspace{0pt}\textit{Image-based}}\\

RAFTStereo \cite{lipson2021raft}  & 10 mixed datasets  & 0.6M & 1.49 & 9.79 & 5.37 & -- & -- & -- & {0.41} & {7.96} & \underline{0.51} \\
IGEVStereo \cite{xu2023iterative}  & SF+M & 0.04M & 1.53 & 9.72 & 5.37 & 1.53 & 13.49 & 2.23 & 0.46 & 9.12 & 0.60 \\
Selective-IGEV \cite{wang2024selective}  & SF+M & 0.04M & 1.46 & 9.33 & 5.46 & 1.64 & 13.78 & 2.14 & 0.48 & 9.66 & 0.60 \\
DEFOM-Stereo \cite{jiang2025defomstereodepthfoundationmodel} & 15 mixed datasets  & 0.7M & 1.29 & 9.12 & 4.25 & -- & -- & -- & \underline{0.39} & \underline{7.08} & \textbf{0.48} \\
MonSter \cite{cheng2025monstermarrymonodepthstereo} & 10 mixed datasets  & 0.6M & 1.25 & 8.65 & 5.19 & -- & -- & -- & 0.41 & 7.73 & 0.60 \\
FoundationStereo \cite{wen2025foundationstereozeroshotstereomatching} & 7 mixed datasets  & 1.3M & 1.78 & 9.82 & 5.40 & -- & -- & -- & 0.40 & 7.20 & 0.57 \\
\rowcolor{gray!20} \multicolumn{12}{l}{\hspace{0pt}\textit{Video-based}}\\
DynamicStereo \cite{karaev2023dynamicstereo} & SF+DR+ISV+\textbf{KD}  & 0.1M & 1.07 & 9.64 & 9.36 & 1.38 & 11.84 & 3.38 & 0.42 & 8.21 & {0.57} \\
BiDAStereo \cite{jing2024matchstereovideos}  & SF+DR+ISV+\textbf{KD}  & 0.1M & \underline{0.90} & \underline{7.90} & \underline{3.92} & \underline{1.33} & \underline{11.34} & \underline{1.97} & 0.42 & 8.33 & 0.58 \\
Stereo Any Video (ours) & SF+DR+ISV+VK & 0.1M & \textbf{0.77} & \textbf{6.52} & \textbf{3.17} & \textbf{0.99} & \textbf{9.09} & \textbf{1.48} & \textbf{0.35} & \textbf{6.05} & 0.61   \\
\bottomrule
\end{tabular}
\vspace{-.9em}
\label{tab:mix}
\end{table*}

\subsection{Benchmark Datasets and Metric}
\noindent \textbf{Dataset.} Our work focuses on videos captured with moving cameras, making commonly used image benchmarks unsuitable, such as Middlebury \cite{middlebury}, ETH3D \cite{eth3d}, and KITTI 2012/2015 \cite{kitti} (two frames per scene). To evaluate our method, we use five synthetic and two real-world stereo video datasets, all featuring dynamic scenes. Sintel \cite{sintel} is a synthetic dataset comprising movie sequences available in both clean and final passes. Spring \cite{Mehl2023_Spring} is a high resolution synthetic dataset. Dynamic Replica \cite{karaev2023dynamicstereo} is a synthetic indoor dataset featuring non-rigid objects, such as animals and people. Infinigen SV \cite{jing2024matchstereovideosbidirectional} is a synthetic dataset focusing on outdoor natural environments. Virtual KITTI2 \cite{cabon2020vkitti2} is a synthetic dataset simulating outdoor driving scenarios. KITTI Depth \cite{kittidepth} is a real-world outdoor dataset captured for autonomous driving applications, with sparse depth maps obtained from a LiDAR sensor. South Kensington SV \cite{jing2024matchstereovideosbidirectional} is a real-world dataset capturing daily scenarios from a stereo camera for qualitative evaluation.

\noindent \textbf{Metric.} For accuracy evaluation, we use End-Point Error (EPE), which measures the average per-pixel disparity error. To assess temporal consistency, we employ Temporal EPE (TEPE) \cite{karaev2023dynamicstereo}, which quantifies the variation in EPE across the temporal dimension. We also compute $\delta_{n\text{-px}}^{t}$, which represents the proportion of pixels with TEPE exceeding $n$. For all metrics, lower values indicate better performance.

\subsection{Implementation Details} \label{Implementation Details}
We implement Stereo Any Video in PyTorch and train on NVIDIA A100 GPUs. The model is trained with a batch size of $8$ on SceneFlow \cite{mayer2016large} for $120k$ iterations and fine-tuned on a mixed dataset for $80k$ iterations consisting of Dynamic Replica \cite{karaev2023dynamicstereo} training set, Infinigen SV \cite{jing2024matchstereovideosbidirectional} training set, and Virtual KITTI2 \cite{cabon2020vkitti2}. Before being fed into the network, sequences are randomly cropped to $256 \times 512$. We use the AdamW optimizer with a learning rate of $2e^{-4}$ and apply a one-cycle learning rate schedule. The full training process takes approximately 6 days. Following \cite{karaev2023dynamicstereo}, we apply various data augmentation techniques, including random cropping, rescaling, and saturation shifts. The sequence length is set to $T=5$ during training, and $T=20$ for evaluation. The number of iterations in GRU is set to $N=10$ in training, and $N=20$ in evaluation.

\begin{figure*}[t]
   \begin{center}
   \includegraphics[width=.88\linewidth]{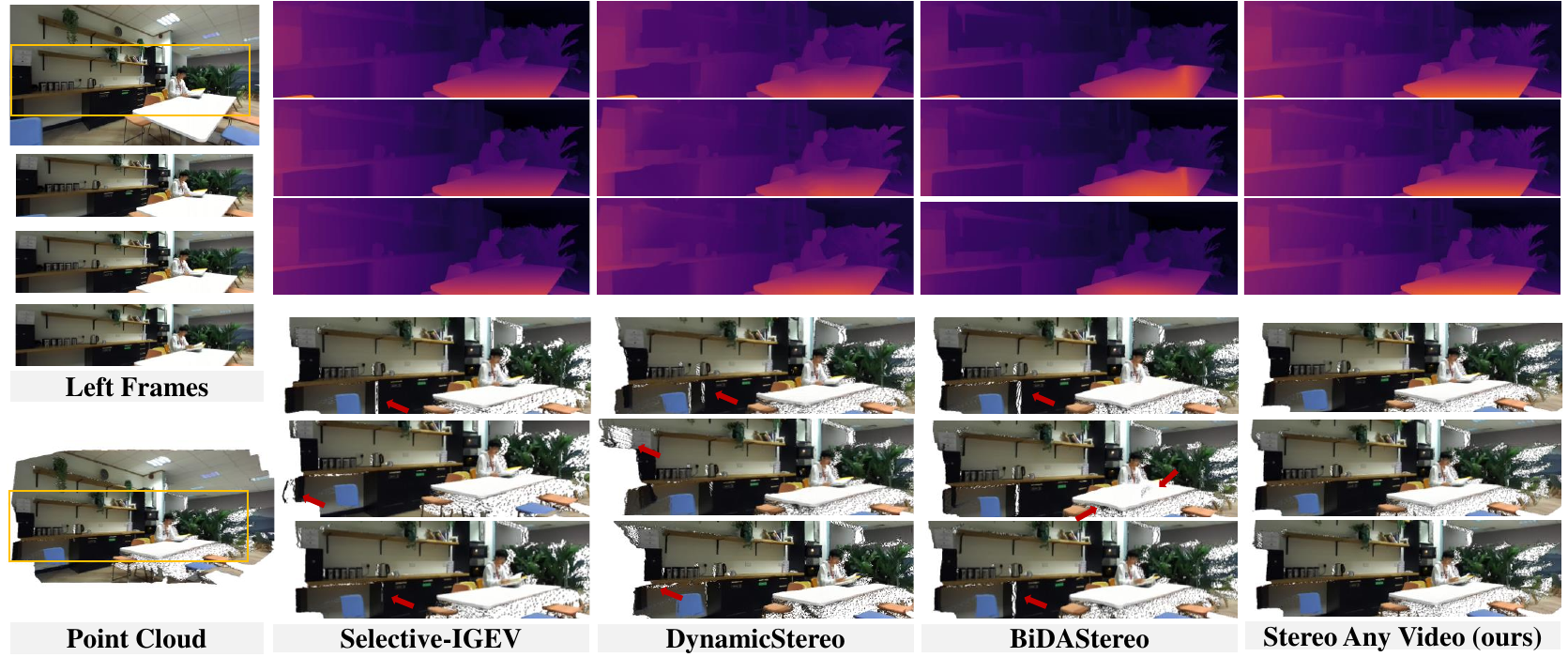}
   \end{center}
   \vspace{-.8em}
    \caption{Qualitative comparison on a dynamic indoor scenario from the South Kensington SV dataset \cite{jing2024matchstereovideosbidirectional}. \textbf{Top:} Input left frames and predicted disparities for the same region across three frames using different methods. \textbf{Bottom:} Disparities converted into globally aligned point clouds and rendered from a camera view for comparison. Failure regions are highlighted with red arrows. Our method produces consistent and accurate disparities without flickering. See video in the supplementary material for better visualization.}
    \vspace{-.5em}
    \label{fig:SK_indoor}
\end{figure*}

\begin{figure*}[t]
   \begin{center}
   \includegraphics[width=.88\linewidth]{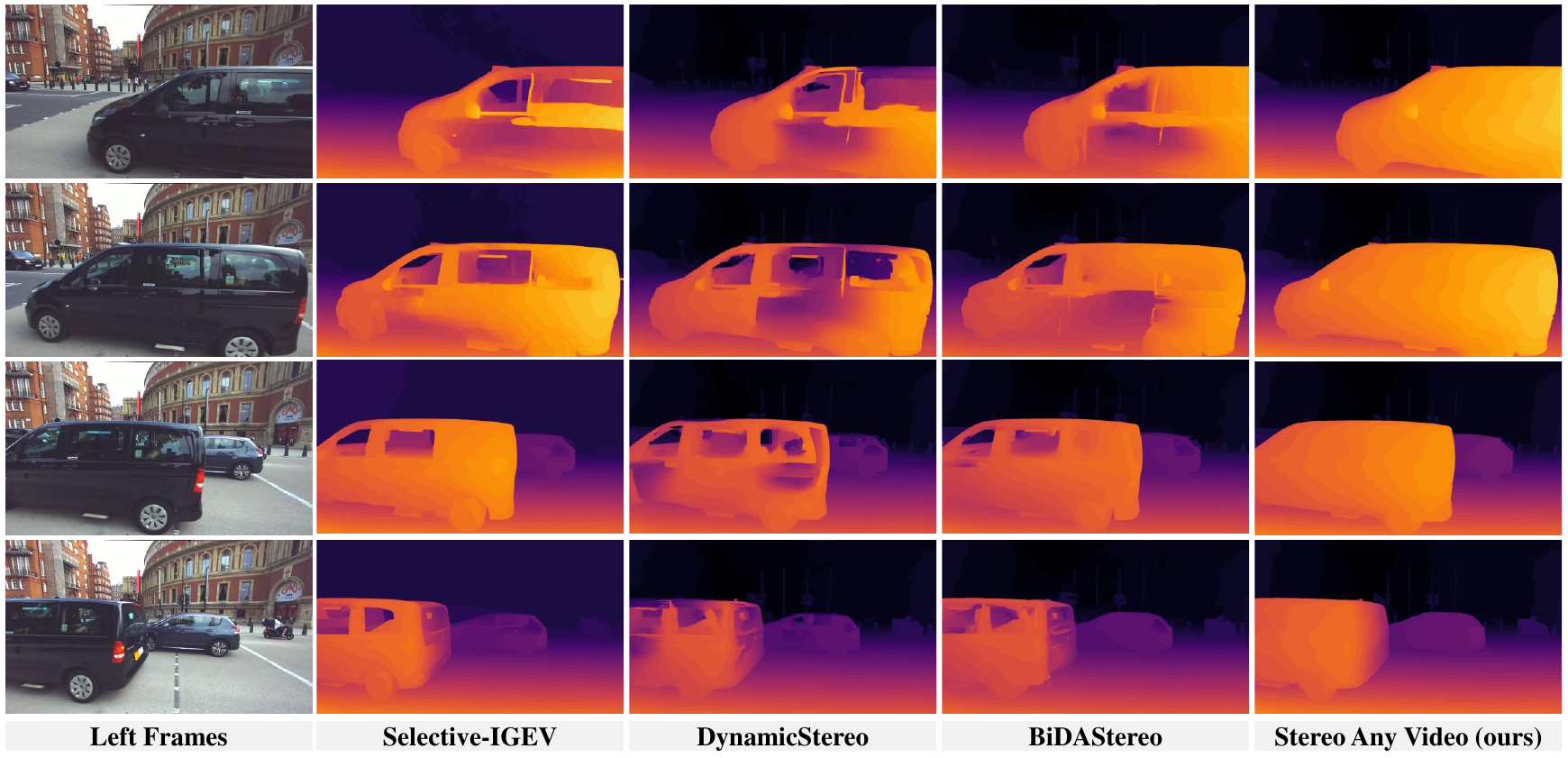}
   \end{center}
   \vspace{-.8em}
    \caption{Qualitative comparison on a dynamic outdoor scenario from the South Kensington SV dataset \cite{jing2024matchstereovideosbidirectional}.} 
    \vspace{-.5em}
    \label{fig:SK_outdoor}
\end{figure*}

\subsection{Evaluation}
\noindent \textbf{Quantitative Results.} \Cref{tab:zeroshot} presents the results when models are trained solely on SceneFlow \cite{mayer2016large} and evaluated on four dynamic synthetic datasets including diverse indoor and outdoor environments. The results clearly demonstrate that our model achieves state-of-the-art zero-shot generalization across nearly all metrics, excelling in both accuracy and temporal consistency. \Cref{tab:mix} reports the results when models are allowed to train on any existing datasets. Notably, despite being trained solely on synthetic datasets, our model surpasses comparison methods that are trained on real-world datasets or share the same domain as the test tests by at least 15\% on diverse benchmarks. Specifically, compared to the robust version of RAFT-Stereo, which is trained on 10 mixed datasets, including in-domain sets such as Sintel \cite{sintel} and KITTI \cite{kitti}, our approach achieves superior performance across most metrics. These results highlight the strong generalization capability of our method when applied to unseen domains.

\noindent \textbf{Qualitative Results.} \Cref{fig:SK_indoor} and \Cref{fig:SK_outdoor} present qualitative results using the checkpoints from \Cref{tab:mix} on dynamic real-world stereo videos from the South Kensington SV dataset \cite{jing2024matchstereovideosbidirectional}. For the indoor scenario shown in \Cref{fig:SK_indoor}, the upper rows display three input left frames along with their predicted disparities, while the subsequent rows visualize the globally aligned reconstructed point cloud. The red arrows highlight that our method produces temporally consistent and accurate disparities, whereas other methods exhibit noticeable artifacts. Specifically, Selective-IGEV and DynamicStereo suffer from flickering artifacts, with cabinets and walls displaying low-frequency oscillations. BiDAStereo fails to maintain an accurate shape, resulting in a distorted table. For the outdoor scenario in \Cref{fig:SK_outdoor}, the driving car presents a challenging case due to its transparent surfaces and rapid motion. As shown, all other methods struggle with this case, failing to produce reliable disparity estimations. In contrast, our method demonstrates superior performance, maintaining accurate and stable predictions.

\begin{table}[tb] 
\caption{Ablation study on the foundation priors, correlation, cost volume encoder, convex upsampling, attention, and inference settings of our model on Sintel \cite{sintel} and Dynamic Replica \cite{karaev2023dynamicstereo}. The approach used in our final model is underlined.}
\vspace{-.6em}
\centering
\setlength{\tabcolsep}{6.pt}
\footnotesize
\begin{tabular}{l|cc|cc}
\toprule
\multirow{2}{*}{\; \; Setting Variations} & \multicolumn{2}{c}{Sintel Final} & \multicolumn{2}{c}{Dynamic Replica}    \\
\cmidrule(lr){2-5}
 & TEPE$\downarrow $ & $\delta_{1px}^{t}\downarrow$ & TEPE$\downarrow $ & $\delta_{1px}^{t}\downarrow$ \\
\midrule
\rowcolor{gray!20} \multicolumn{5}{l}{\hspace{0pt}\textit{Foundation Priors}}\\
w/o Depth   & 1.47 & 12.35 & 0.092 & 1.05 \\
DepthAnythingV2-S \cite{yang2024depth} & 1.33 & 11.27 & 0.088 & 0.99 \\
\underline{VDA-S \cite{chen2025videodepthanythingconsistent}}  & 1.27 & \textbf{10.98} & \textbf{0.083} & \textbf{0.78}  \\
VDA-L \cite{chen2025videodepthanythingconsistent}              & \textbf{1.20} & 12.47 & \textbf{0.083} & 0.82  \\
\rowcolor{gray!20} \multicolumn{5}{l}{\hspace{0pt}\textit{Correlation}}\\
1D + 2D local  \cite{li2022practical}  & 1.27 & \textbf{10.98} & 0.083 & 0.78  \\
1D all-to-all-pairs                    & 1.23 & 11.36 & 0.079 & 0.78  \\
2D all-to-all-pairs                    & 1.29 & 11.95 & 0.085 & 0.77  \\
3D all-to-all-pairs                    & 4.51 & 22.90 & 1.313 & 2.01  \\
\underline{1D + 2D all-to-all-pairs}   & \textbf{1.21} & 11.33 & \textbf{0.076} & \textbf{0.75}  \\
\rowcolor{gray!20} \multicolumn{5}{l}{\hspace{0pt}\textit{Cost Volume Encoder}}\\
Convolution                    & 1.26 & 11.54 & 0.081 & 0.77  \\
\underline{MLP + Convolution}  & \textbf{1.21} & \textbf{11.33} & \textbf{0.076} & \textbf{0.75}  \\
\rowcolor{gray!20} \multicolumn{5}{l}{\hspace{0pt}\textit{Convex Upsampling}}\\
Bilinear                & 1.26 & 12.74 & 0.085 & 1.05 \\
Convex \cite{teed2020raft}              & 1.21 & 11.33 & 0.076 & \textbf{0.75}  \\
\underline{Temporal Convex}   & \textbf{1.04} & \textbf{10.21} & \textbf{0.067} & 0.79 \\
\rowcolor{gray!20} \multicolumn{5}{l}{\hspace{0pt}\textit{Attention in GRU}}\\
w/o Attention & \textbf{1.04} & {10.21} & {0.067} & 0.79 \\
Temporal & {1.07} & 10.23 & 0.064 & 0.79 \\
\underline{Temporal + Spatial} & {1.07} & \textbf{9.98} & \textbf{0.057} & \textbf{0.67} \\
\rowcolor{gray!20} \multicolumn{5}{l}{\hspace{0pt}\textit{Inference Stage}}\\
\underline{$\frac{1}{16}\rightarrow\frac{1}{8}\rightarrow\frac{1}{4}$} & \textbf{1.07} & \textbf{9.98} & \textbf{0.057} & \textbf{0.67} \\
$\frac{1}{32}\rightarrow\frac{1}{16}\rightarrow\frac{1}{8}\rightarrow\frac{1}{4}$   & 1.08 & 10.82 & 0.058 & 0.70 \\
\bottomrule
\end{tabular}
\label{tab:ablation}
\vspace{-1.9em}
\end{table}

\subsection{Ablation Study}
\noindent \textbf{Foundation Priors.} As shown in \Cref{tab:ablation}, incorporating foundation priors from a monocular depth model improves performance. Notably, VDA \cite{chen2025videodepthanythingconsistent} outperforms DepthAnythingV2 \cite{yang2024depth} due to its stronger temporal consistency. Interestingly, increasing the backbone size does not yield additional performance gains. To investigate this, we conducted further experiments (\Cref{tab:vda}) by varying depth and CNN feature channels based on VDA-L. However, the results remained similar. According to \cite{chen2025videodepthanythingconsistent}, VDA-L has $13\times$ more parameters than VDA-S (381.8M vs. 28.4M), yet the performance gain is marginal ($\delta_1(\uparrow)$: 0.944 vs. 0.942 on KITTI; 0.971 vs. 0.959 on NYUv2). This suggests that the larger backbone does not provide richer priors. Moreover, large backbone requires additional CNN channels and parameters for adaptation, increasing training complexity. Therefore, we adopt VDA-S in the final model.

\noindent \textbf{Correlation.} To evaluate the effectiveness of our proposed all-to-all-pairs correlation, we replace it with local correlation \cite{li2022practical} and experiment with using only 1D or 2D correlation instead of alternating between them. As shown in \Cref{tab:ablation}, all these modifications degrade performance. Notably, constructing 3D correlations—where the search window spans height, width, and time—further worsens results. This suggests that frames cannot be effectively utilized without proper alignment \cite{jing2024matchstereovideosbidirectional}.

\noindent \textbf{Cost Volume Encoder.} Instead of directly encoding cost volumes with convolutions, we find that introducing an MLP layer to extract a compact representation before applying convolutions enhances cost aggregation, leading to improved final results.

\noindent \textbf{Convex Upsampling.} We compare bilinear upsampling and general convex upsampling \cite{teed2020raft} to our proposed temporal convex upsampling module. As observed, our approach achieves better temporal consistency by incorporating learnable weights with adjacent frames.

\noindent \textbf{Attention in GRU.} We further investigate different attention mechanisms within the GRU update block. As shown in \Cref{tab:ablation}, combining temporal and spatial attention together leads to the best performance.

\noindent \textbf{Inference Stage.} We also explore inference starting at different resolutions. As can be seen in \Cref{tab:ablation}, a three-stage inference setup already achieves strong results.

\subsection{Application}
\begin{figure}[t]
   \begin{center}
   \includegraphics[width=.95\linewidth]{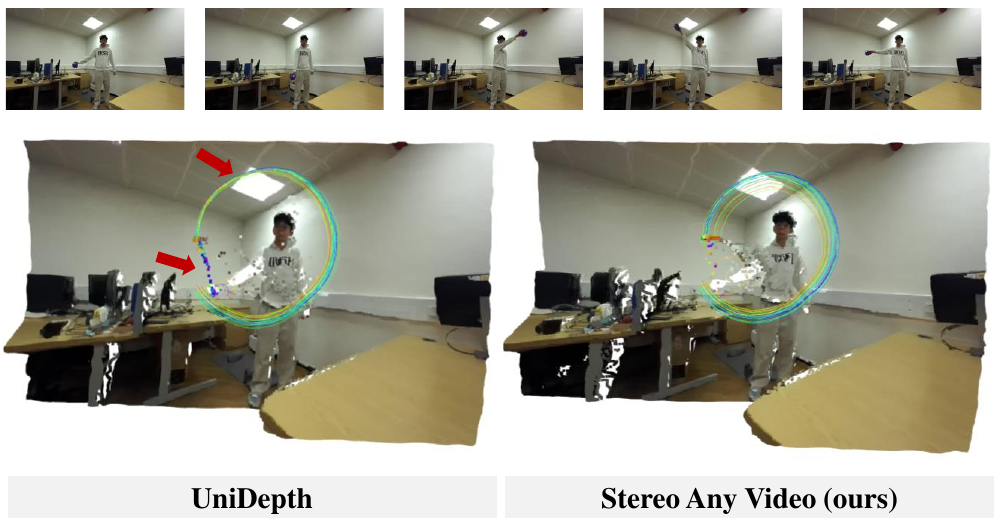}
   \end{center}
   \vspace{-1.3em}
    \caption{Visualization of 3D point tracking based on metric depth generated from UniDepth and our method.} 
    \vspace{-.3em}
    \label{fig:application}
\end{figure}

Our method enhances downstream applications that rely on accurate metric depth, such as 3D point tracking. As shown in \Cref{fig:application}, we visualize results using the state-of-the-art 3D point tracking method DELTA \cite{ngo2025deltadenseefficientlongrange}.  Our predicted metric depth yields more points with precise trajectories than UniDepth \cite{piccinelli2024unidepthuniversalmonocularmetric}, which gives less missing points as highlighted by the red arrows. Additional application, such as visual effects, are available in the supplementary material.


\subsection{Limitation and Future Work}
We report the inference efficiency results on a single NVIDIA A6000 GPU in \Cref{tab:efficiency}. As shown, due to the use of foundation priors, our method incurs relatively high computational and memory costs. Looking ahead, we aim to develop a model zoo for this method, including both a large and a light version.

\begin{table}[tb] 
\caption{Ablation study on the effect of different channel configurations for depth features and CNN features, based on VDA-L \cite{chen2025videodepthanythingconsistent}, on Sintel \cite{sintel}. Trainable parameters are also reported.}
\vspace{-.6em}
\centering
\setlength{\tabcolsep}{8.pt}
\footnotesize
\begin{tabular}{c|cc|c}
\toprule
{Depth \& CNN} & \multicolumn{2}{c}{Sintel Final} & \multirow{2}{*}{Para. (M)} \\
\cmidrule(lr){2-3}
Channel Number & TEPE$\downarrow $ & $\delta_{1px}^{t}\downarrow$ & \\
\midrule
\rowcolor{gray!20} \multicolumn{4}{l}{\hspace{0pt}\textit{VDA-L} \cite{chen2025videodepthanythingconsistent}}\\
(32, 96)    & 1.20 & 12.47 & 8.4  \\
(128, 64)    & 1.58 & 14.52 & 13.3  \\
(128, 128)    & 1.31 & 13.21 &  19.3 \\
\bottomrule
\end{tabular}
\label{tab:vda}
\vspace{-1em}
\end{table}

\begin{table}[t]
\footnotesize
   \begin{center}
       \caption{Comparison of different methods on model trainable [non-trainable] parameters, inference GPU memory, and multiply-accumulate (MAC) on 20 frames with resolution $720\times$1280.}
       \vspace{-.5em}
     \begin{tabular}{m{3.0cm}<{\centering}|m{1.2cm}<{\centering}m{1.2cm}<{\centering}m{1.2cm}<{\centering}}
        \toprule
        Methods &  Para. (M) & Mem. (G) & MACs (T) \\
        \midrule
        DynamicStereo \cite{karaev2023dynamicstereo} & 20.5 & 35.1 & 182.3 \\
        BiDAStereo \cite{jing2024matchstereovideos}  & 12.2 & 41.1 & 186.6   \\
        Stereo Any Video (ours)  & 9.4 [28.4] & 41.1 & 303.4  \\
        \bottomrule
     \end{tabular}
\label{tab:efficiency}
\vspace{-1.5em}
   \end{center}
\end{table}

\section{Conclusion}
In this paper, we introduce Stereo Any Video, a robust framework for video stereo matching. By leveraging rich monocular priors, an all-to-all-pairs correlation mechanism, and a temporal convex upsampling layer, our approach produces temporally consistent and accurate disparities without relying on additional inputs such as camera pose or optical flow. Extensive evaluations show that our model achieves state-of-the-art performance on multiple standard benchmarks under zero-shot settings. Furthermore, our method generalizes well to real-world stereo videos, handling diverse content, motion, and camera movements effectively.

{
    \small
    \bibliographystyle{ieeenat_fullname}
    \bibliography{main}
}

\clearpage

\appendix

\renewcommand*{\thesection}{\Alph{section}}
\newcommand{\multiref}[2]{\cref{#1}--\ref{#2}}
\renewcommand{\thetable}{S\arabic{table}}
\renewcommand{\thefigure}{S\arabic{figure}}

\setcounter{table}{0}
\setcounter{figure}{0}
\setcounter{page}{1}

\maketitlesupplementary

In this supplementary material, we provide additional details and more qualitative results. We highly recommend referring to the \textbf{video} we provided, since the visual quality of the disparities can be better accessed with videos.

\section{Settings on Table 1.}
We compare the temporal consistency of video disparity estimated by DepthCrafter \cite{hu2024-DepthCrafter} and RAFTStereo \cite{lipson2021raft} by collecting ratings from 10 recruited participants. Each participant's rating is determined by the average of two scores: flickering region \( f_r \) and flickering strength \( f_s \) out of three possible levels: \( 0 \), \( 1 \), and \( 2 \). A higher \( f_r \) indicates a larger flickering region within the video disparity, while a higher \( f_s \) reflects a greater depth discrepancy between video frames.

\section{Feature Extraction}
In the feature extraction stage, when processing a video sequence with the monocular video depth model, we first resize it to ensure its dimensions are divisible by 14, maintaining consistency with the model's pretrained patch size. After obtaining the feature maps, we resize the image back to its original dimensions. Unlike previous methods \cite{lipson2021raft, li2022practical}, which normalize the image directly to the range [-1,1], we apply mean and standard deviation normalization based on ImageNet pre-trained models. This ensures better alignment with the VDA framework. The monocular depth model produces feature maps with 32 channels, while the CNN encoders extract both image and context features with 96 channels each. These feature maps are concatenated to form a 128-channel representation, which serves as input to the subsequent correlation module.

\begin{figure}[t]
\begin{lstlisting}[language=Python, 
                   frame=tb,
                   xleftmargin=0pt,
                   xrightmargin=0pt,
                   resetmargins=true,
                   breaklines=true,
                   breakindent=0pt,
                   columns=fullflexible,
                   keepspaces=true]
# F[n, 2, t, h, w]      - image flow field
# M[n, c, t, h, w]      - learnable weights
# r                     - upsampling factor

# reshape and normalize mask weights into
# M[n, 1, 27, 1, r, r, t, h, w]
M = softmax(reshape(M), dim=2)

# Patch and reshape the scaled flow
# F[n, 2, 27, 1, 1, 1, t, h, w]
F = unfold_3d(r*F, kernel=[3,3,3])

# Compute the weighted (convex) combination
# F_upsampled[n, 2, 1, r, r, t, h, w]
F_upsampled = sum(M*F, dim=2)

# permute and upsample the flow
# F_upsampled[n, 2, t, r*h, r*w]
F_upsampled = reshape(permute(F_upsampled))

return F_upsampled
\end{lstlisting}
\caption{Pythonic pseudo-code for the implementation of temporal convex upsampling with a kernel size of $3\times3\times3$.}
\label{fig:convex-upsample}
\end{figure}

\section{Temporal Convex Upsampling}
We develop temporal convex upsampling that extends traditional spatial upsampling techniques to the temporal dimension, enabling precise disparity interpolation across both spatial and temporal dimensions. As shown in \Cref{fig:convex-upsample}, our implementation leverages a convex combination approach, which ensures that the upsampled disparity maintains physical consistency while preserving intricate motion patterns. The method operates by first reshaping and normalizing mask weights through a softmax operation, creating a probability distribution across neighboring elements. Subsequently, we extract local patches from the input disparity field using a 3D unfold operation typically with a kernel size of 3, scaling the disparity vectors by the upsampling factor to maintain velocity magnitudes. The core of the algorithm lies in computing the weighted sum of these patches using the normalized mask weights, effectively performing a learned interpolation that respects the underlying motion structure. Our implementation can efficiently handle arbitrary batch sizes and integrate with existing deep learning architectures, making it suitable for video understanding tasks with high-resolution disparity fields.

\section{Datasets}

\subsection{SceneFlow (SF)}
SceneFlow \cite{mayer2016large} consists of three subsets: {FlyingThings3D}, {Driving}, and {Monkaa}.
\begin{itemize}
    \item {FlyingThings3D} is an abstract dataset featuring moving shapes against colorful backgrounds. It contains 2,250 sequences, each spanning 10 frames.
    \item {Driving} includes 16 sequences depicting driving scenarios, with each sequence containing between 300 and 800 frames.
    \item {Monkaa} comprises 48 sequences set in cartoon-like environments, with frame counts ranging from 91 to 501.
\end{itemize}

\subsection{Sintel}
Sintel \cite{sintel} is generated from computer-animated films. It consists of 23 sequences available in both {clean} and {final} rendering passes. Each sequence contains 20 to 50 frames. We use the full sequences of Sintel for evaluation.

\subsection{Dynamic Replica}
Dynamic Replica \cite{karaev2023dynamicstereo} is designed with longer sequences and the presence of non-rigid objects such as animals and humans. The dataset includes:
\begin{itemize}
    \item {484 training sequences}, each with 300 frames.
    \item {20 validation sequences}, each with 300 frames.
    \item {20 test sequences}, each with 900 frames.
\end{itemize}
Following prior methods \cite{karaev2023dynamicstereo, jing2024matchstereovideosbidirectional}, we use the entire training set for model training and evaluate on the first 150 frames of the test set.

\subsection{Infinigen SV}
Infinigen SV \cite{jing2024matchstereovideosbidirectional} is a synthetic dataset designed for outdoor natural environments. It consists of {226 photorealistic videos}, each lasting between 3 and 20 seconds, recorded at 24 fps. The dataset is divided into:
\begin{itemize}
    \item {186 training videos}
    \item {10 validation videos}
    \item {30 testing videos}
\end{itemize}
Similar to Dynamic Replica, we use the full training set for training and evaluate on the first 150 frames of the test set.

\subsection{Virtual KITTI2}
Virtual KITTI2 \cite{cabon2020vkitti2} is a synthetic dataset that simulates outdoor driving scenarios. It consists of five sequence clones from the {KITTI tracking benchmark}, with variations in weather conditions (e.g., fog, rain) and camera configurations.

Since this dataset is being used for video stereo matching evaluation for the first time, we randomly select {10\% of the dataset} as the test set. The selected test sequences are:
\begin{itemize}
    \item \texttt{Scene01\_15-deg-left}
    \item \texttt{Scene02\_30-deg-right}
    \item \texttt{Scene06\_fog}
    \item \texttt{Scene18\_morning}
    \item \texttt{Scene20\_rain}
\end{itemize}

\subsection{KITTI Depth}
KITTI Depth \cite{kittidepth} is a real-world outdoor dataset collected for autonomous driving applications. It provides {sparse depth maps} captured using a LiDAR sensor. Following prior work \cite{li2023temporally}, we use the following test sequences:

\begin{itemize}
    \item \texttt{2011\_09\_26\_drive\_0002\_sync}
    \item \texttt{2011\_09\_26\_drive\_0005\_sync}
    \item \texttt{2011\_09\_26\_drive\_0013\_sync}
    \item \texttt{2011\_09\_26\_drive\_0020\_sync}
    \item \texttt{2011\_09\_26\_drive\_0023\_sync}
    \item \texttt{2011\_09\_26\_drive\_0036\_sync}
    \item \texttt{2011\_09\_26\_drive\_0079\_sync}
    \item \texttt{2011\_09\_26\_drive\_0095\_sync}
    \item \texttt{2011\_09\_26\_drive\_0113\_sync}
    \item \texttt{2011\_09\_28\_drive\_0037\_sync}
    \item \texttt{2011\_09\_29\_drive\_0026\_sync}
    \item \texttt{2011\_09\_30\_drive\_0016\_sync}
    \item \texttt{2011\_10\_03\_drive\_0047\_sync}
\end{itemize}

\subsection{South Kensington SV}
South Kensington SV \cite{jing2024matchstereovideosbidirectional} is a real-world stereo dataset capturing {daily life scenarios} for qualitative evaluation. It consists of {264 stereo videos}, each lasting between {10 and 70 seconds}, recorded at {1280×720 resolution} and {30 fps}. We conduct qualitative evaluations on this dataset.

\section{Comparison Methods}
We compare our approach against representative {image-based stereo matching methods}—RAFTStereo \cite{teed2020raft}, IGEVStereo \cite{xu2023iterative}, and Selective-IGEV \cite{wang2024selective}—as well as {video-based stereo matching methods}—DynamicStereo \cite{karaev2023dynamicstereo} and BiDAStereo \cite{jing2024matchstereovideosbidirectional}. For the evaluations presented in {Table~2}, we use the official model checkpoints provided in the open-source implementations. For {Table~3}, we employ the following specific checkpoints:
\begin{itemize}
    \item {RAFTStereo}: Robust Vision Challenge checkpoint
    \item {IGEVStereo \& Selective-IGEV}: Middlebury fine-tuned checkpoints
    \item {DynamicStereo \& BiDAStereo}: Mixed-dataset fine-tuned checkpoints from \cite{jing2024matchstereovideosbidirectional}
\end{itemize}

\section{Application}
\begin{figure*}[t]
   \begin{center}
   \includegraphics[width=1\linewidth]{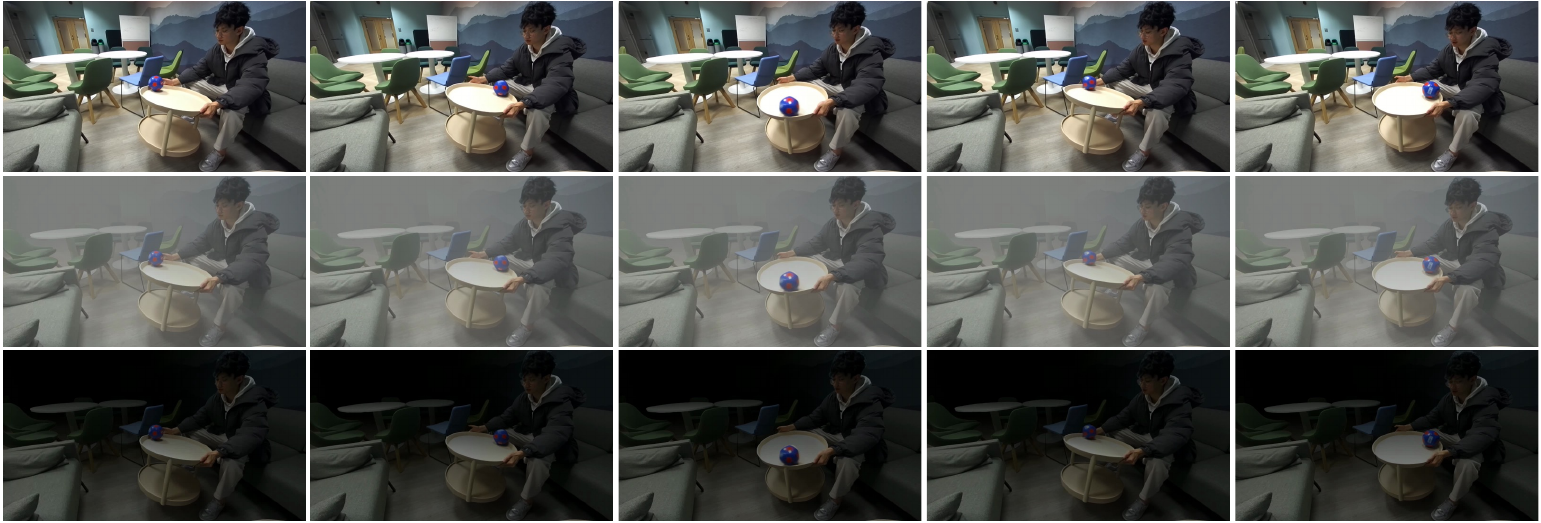}
   \end{center}
   \vspace{-1.2em}
    \caption{Examples of visual effects that could benefit from using our method, including adding fog effects and adjusting light conditions.} 
    \vspace{-.5em}
    \label{fig:sup_visual_effect}
\end{figure*}

\begin{figure*}[t]
   \begin{center}
   \includegraphics[width=1\linewidth]{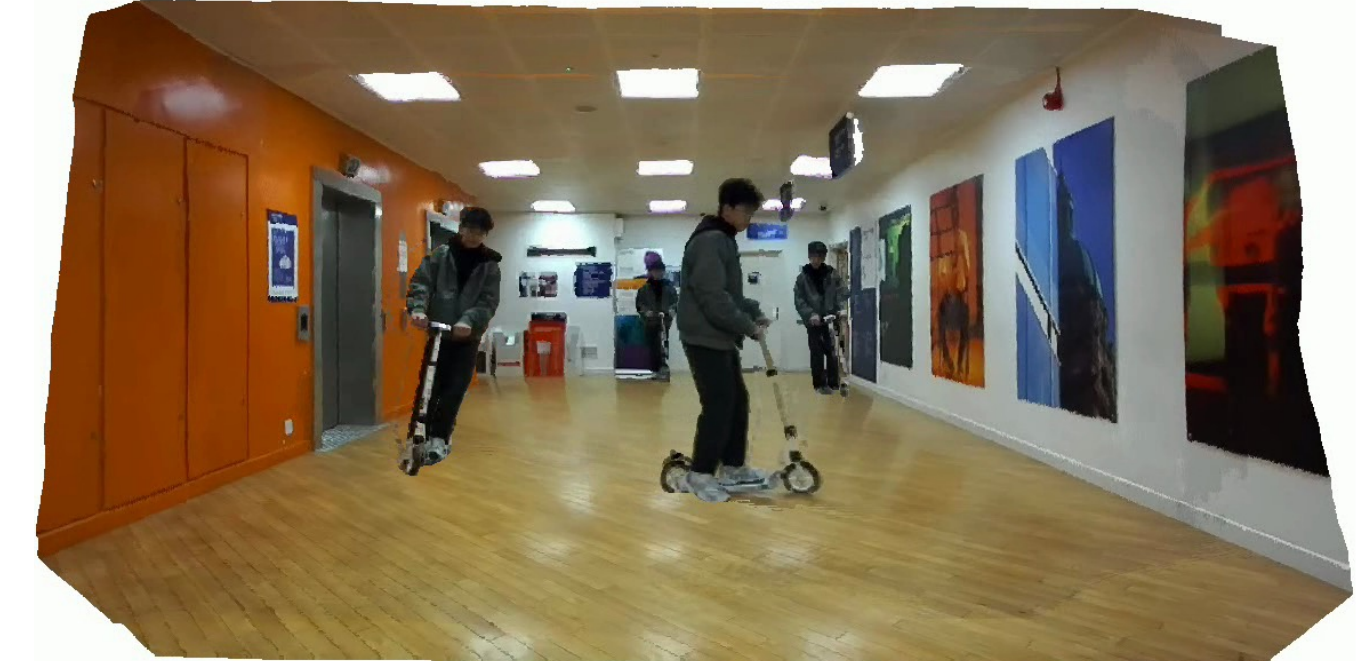}
   \end{center}
   \vspace{-1.2em}
    \caption{Another demo prediction on a dynamic real-world stereo video using our method.} 
    \vspace{-.5em}
    \label{fig:sup_demo}
\end{figure*}

\Cref{fig:sup_visual_effect} demonstrates additional applications that benefit from our method's ability to produce both accurate and temporally consistent depth sequences, including adding atmospheric fog effects and the adjustment of lighting conditions. Specifically, we implement these effects by blending the input video frames with supplementary color maps, with the blending parameters determined by the estimated depth values to simulate varying transparency at the pixel level. As illustrated in the figure, the resulting frames exhibit high consistency in color without perceptible flickering, further corroborating the robust temporal consistency achieved by our method.

\section{Qualitative Results on Real-world Datasets}

\begin{figure*}[t]
   \begin{center}
   \includegraphics[width=1\linewidth]{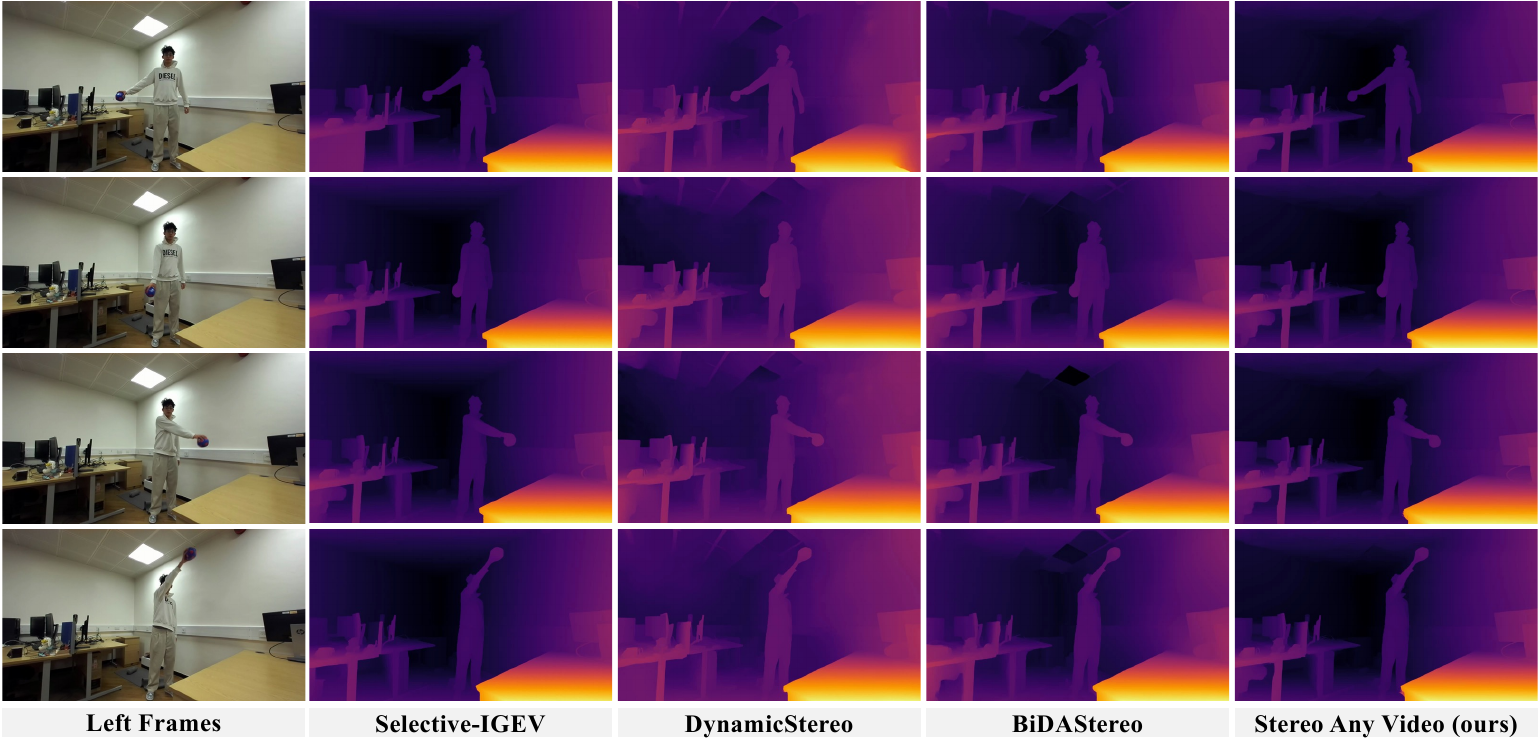}
   \end{center}
   \vspace{-1.2em}
    \caption{Qualitative comparison on a dynamic indoor scenario from the South Kensington SV dataset \cite{jing2024matchstereovideosbidirectional}.} 
    \vspace{-.5em}
    \label{fig:sup_indoor_031}
\end{figure*}

\begin{figure*}[t]
   \begin{center}
   \includegraphics[width=1\linewidth]{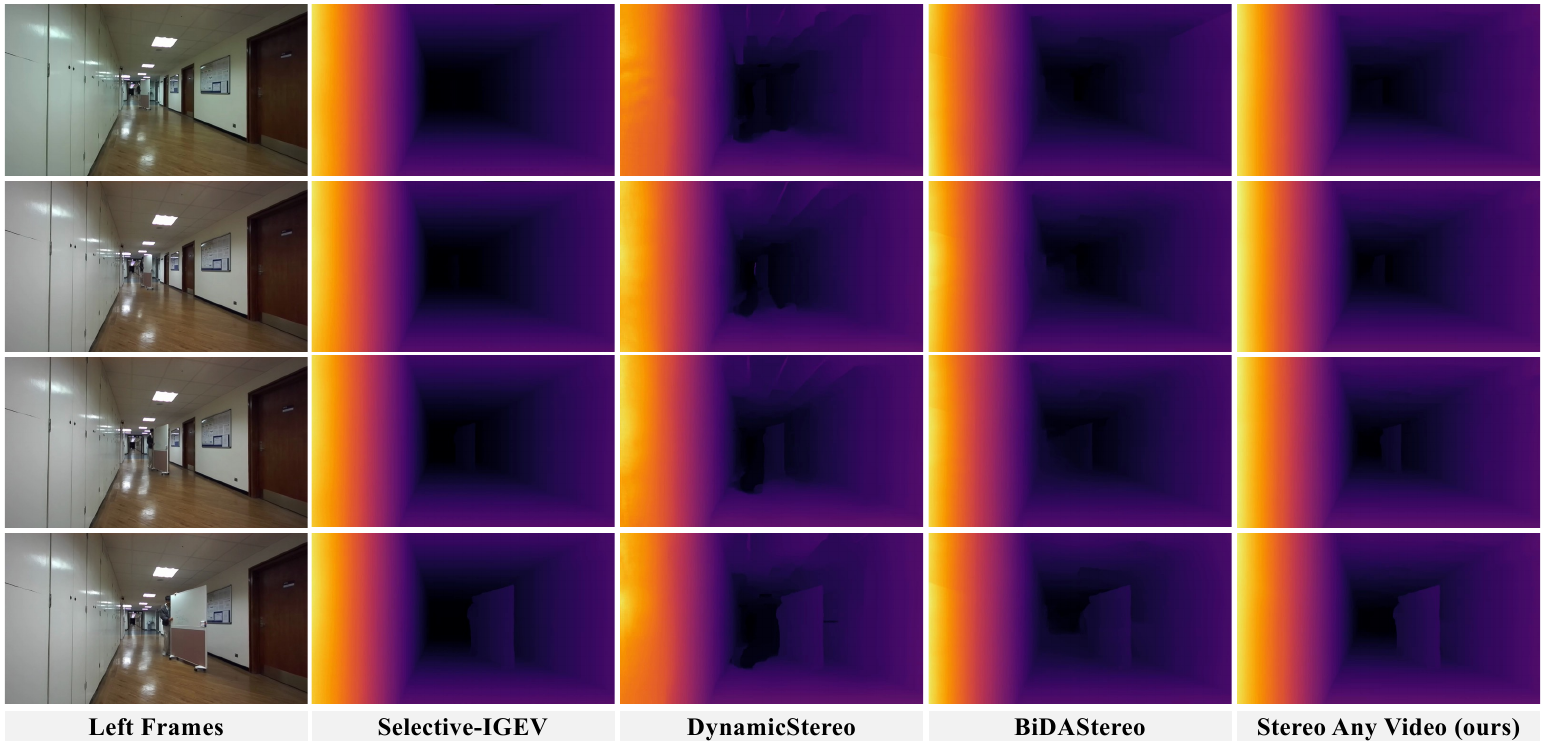}
   \end{center}
   \vspace{-1.2em}
    \caption{Qualitative comparison on a dynamic indoor scenario from the South Kensington SV dataset \cite{jing2024matchstereovideosbidirectional}.} 
    \vspace{-.5em}
    \label{fig:sup_indoor_039}
\end{figure*}

\begin{figure*}[t]
   \begin{center}
   \includegraphics[width=1\linewidth]{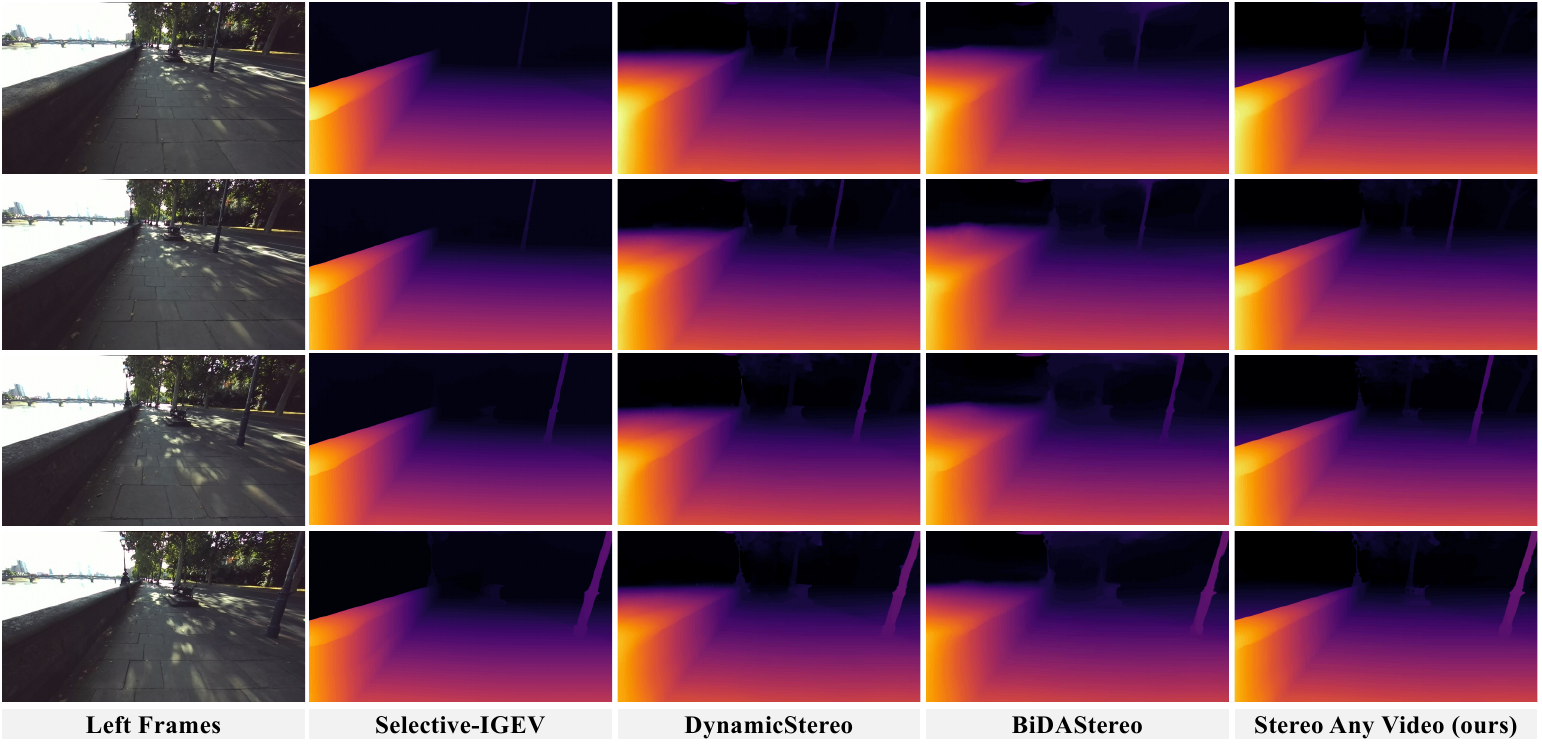}
   \end{center}
   \vspace{-1.2em}
    \caption{Qualitative comparison on a dynamic outdoor scenario from the South Kensington SV dataset \cite{jing2024matchstereovideosbidirectional}.} 
    \vspace{-.5em}
    \label{fig:sup_outdoor_007}
\end{figure*}

\begin{figure*}[t]
   \begin{center}
   \includegraphics[width=1\linewidth]{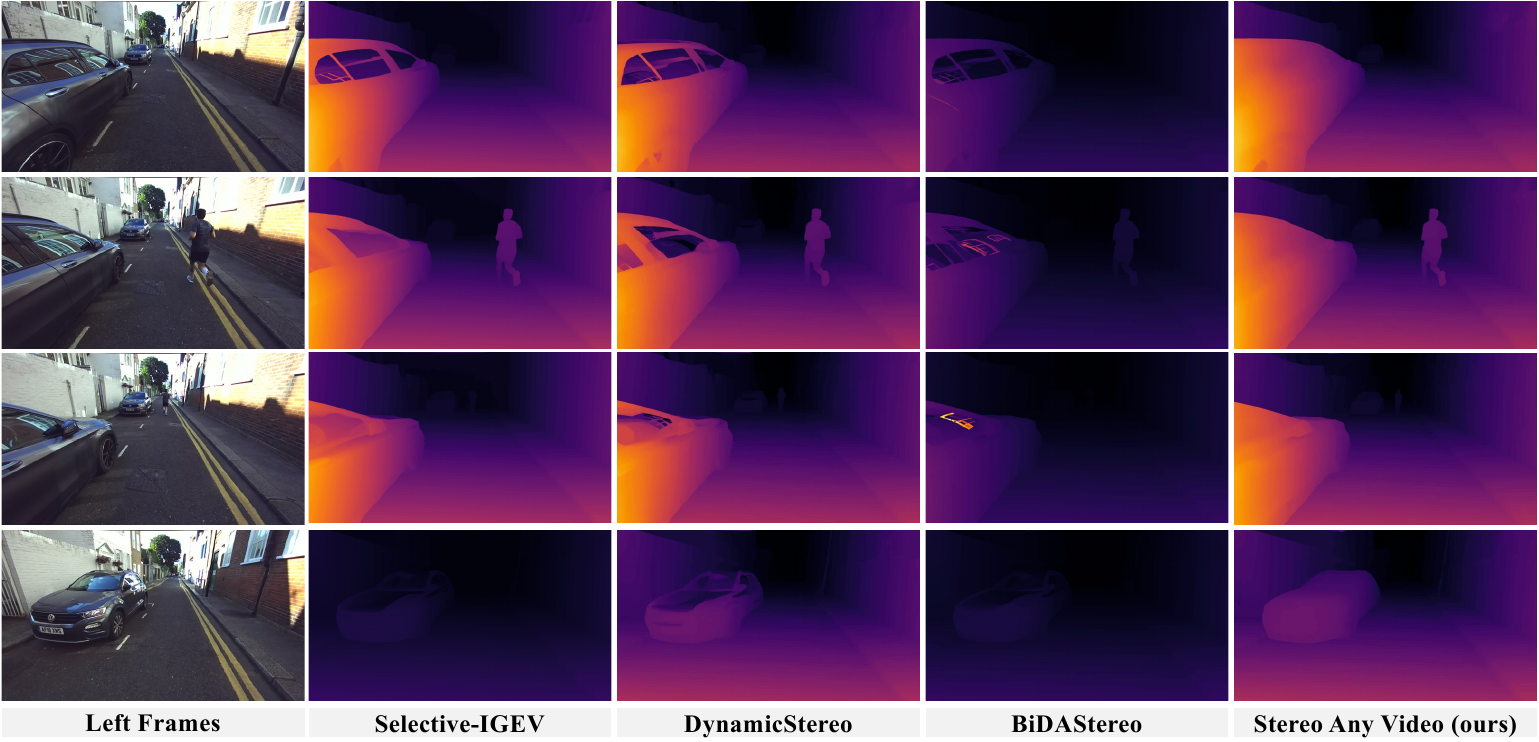}
   \end{center}
   \vspace{-1.2em}
    \caption{Qualitative comparison on a dynamic outdoor scenario from the South Kensington SV dataset \cite{jing2024matchstereovideosbidirectional}.} 
    \vspace{-.5em}
    \label{fig:sup_outdoor_026}
\end{figure*}

\begin{figure*}[t]
   \begin{center}
   \includegraphics[width=1\linewidth]{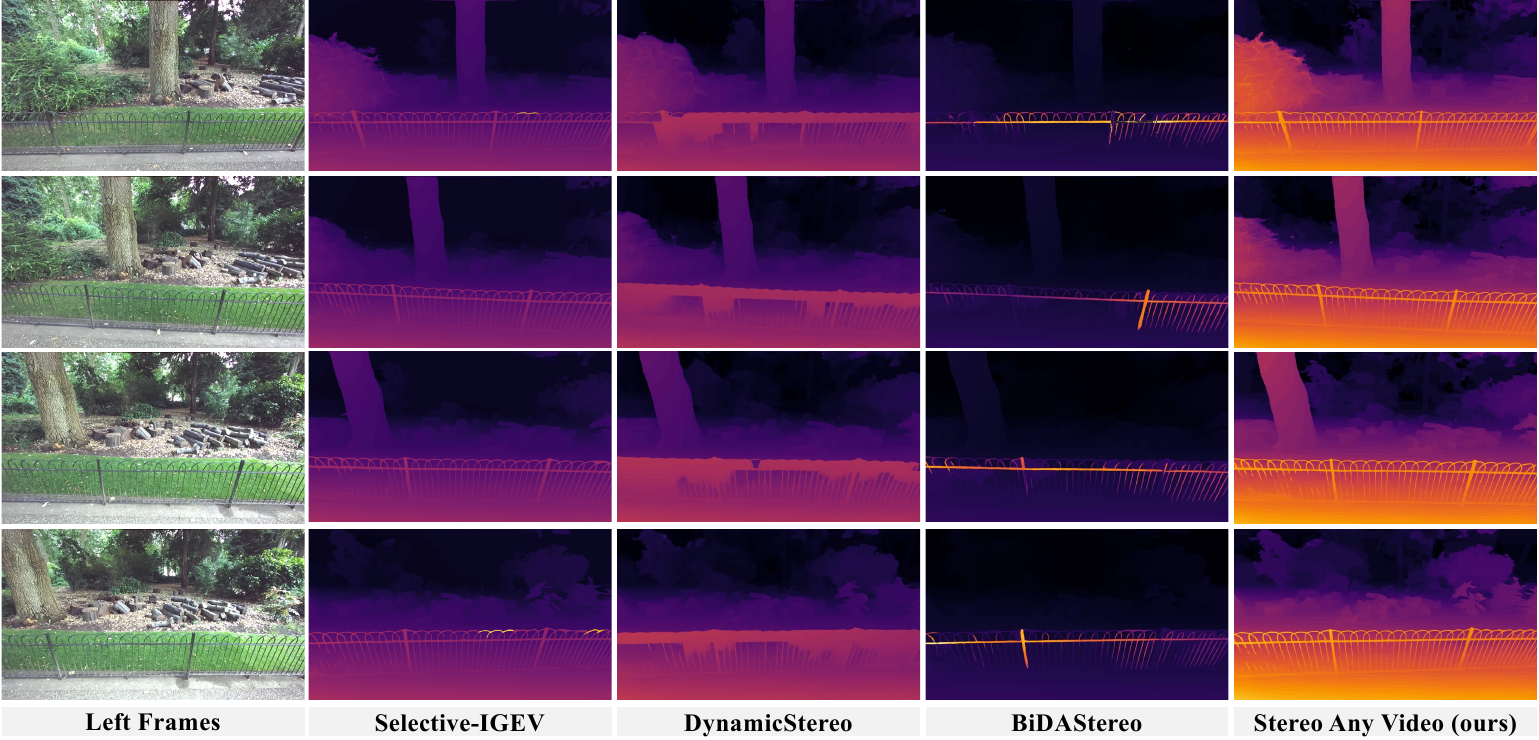}
   \end{center}
   \vspace{-1.2em}
    \caption{Qualitative comparison on a dynamic outdoor scenario from the South Kensington SV dataset \cite{jing2024matchstereovideosbidirectional}.} 
    \vspace{-.5em}
    \label{fig:sup_outdoor_058}
\end{figure*}

\Cref{fig:sup_demo} gives another demo on a dynamic real world video predicted using our method. \Cref{fig:sup_indoor_031} and \Cref{fig:sup_indoor_039} demonstrate comparison on real world indoor scenes. \Cref{fig:sup_outdoor_007}, \Cref{fig:sup_outdoor_026}, and \Cref{fig:sup_outdoor_058} give more examples on real world outdoor scenes.

\section{Qualitative Results on Synthetic Datasets}

\begin{figure*}[t]
   \begin{center}
   \includegraphics[width=1\linewidth]{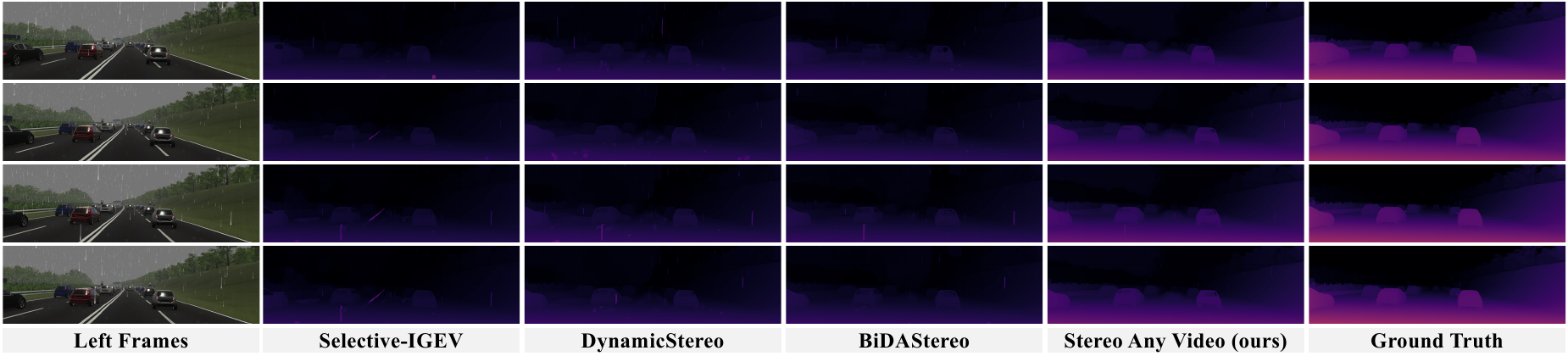}
   \end{center}
   \vspace{-1.2em}
    \caption{Qualitative comparison on Virtual KITTI2 dataset \cite{cabon2020vkitti2}.} 
    \vspace{-.5em}
    \label{fig:sup_Scene20}
\end{figure*}

\begin{figure*}[t]
   \begin{center}
   \includegraphics[width=1\linewidth]{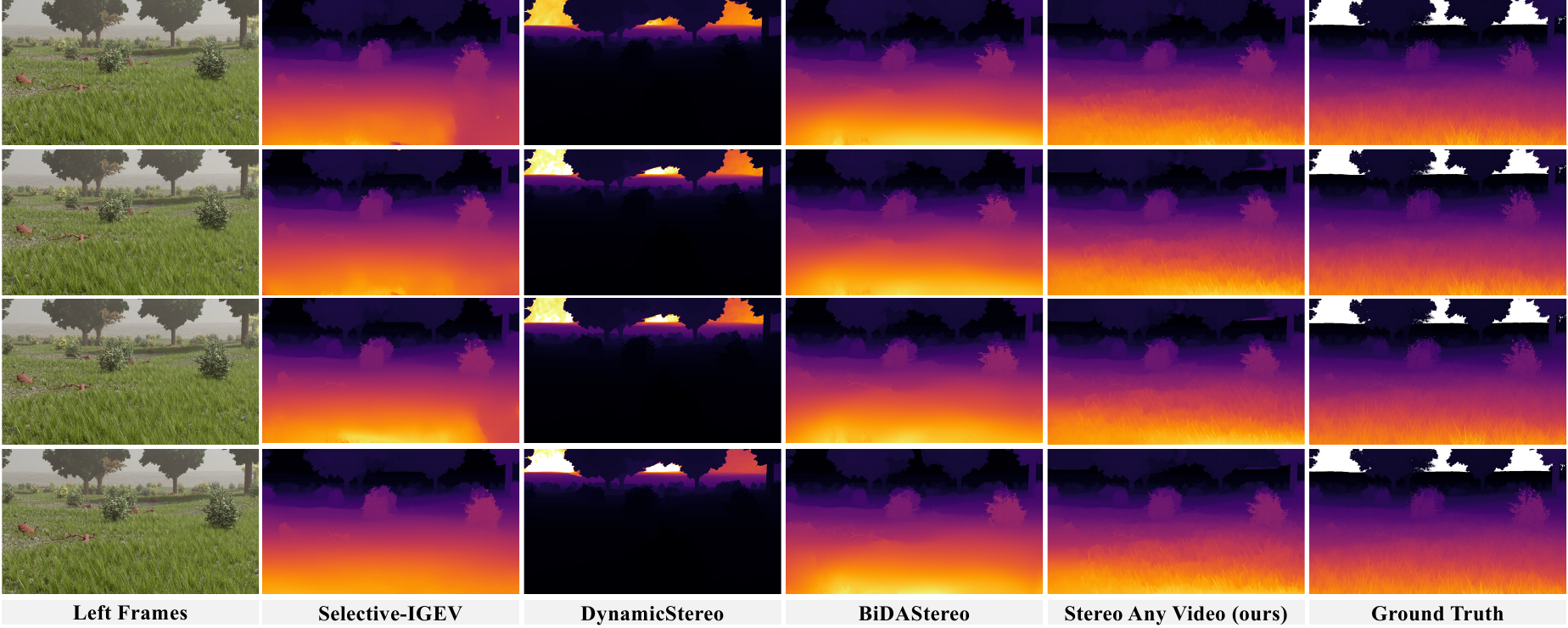}
   \end{center}
   \vspace{-1.2em}
    \caption{Qualitative comparison on Infinigen SV dataset \cite{jing2024matchstereovideosbidirectional}.} 
    \vspace{-.5em}
    \label{fig:sup_1af03d8d}
\end{figure*}

\begin{figure*}[t]
   \begin{center}
   \includegraphics[width=1\linewidth]{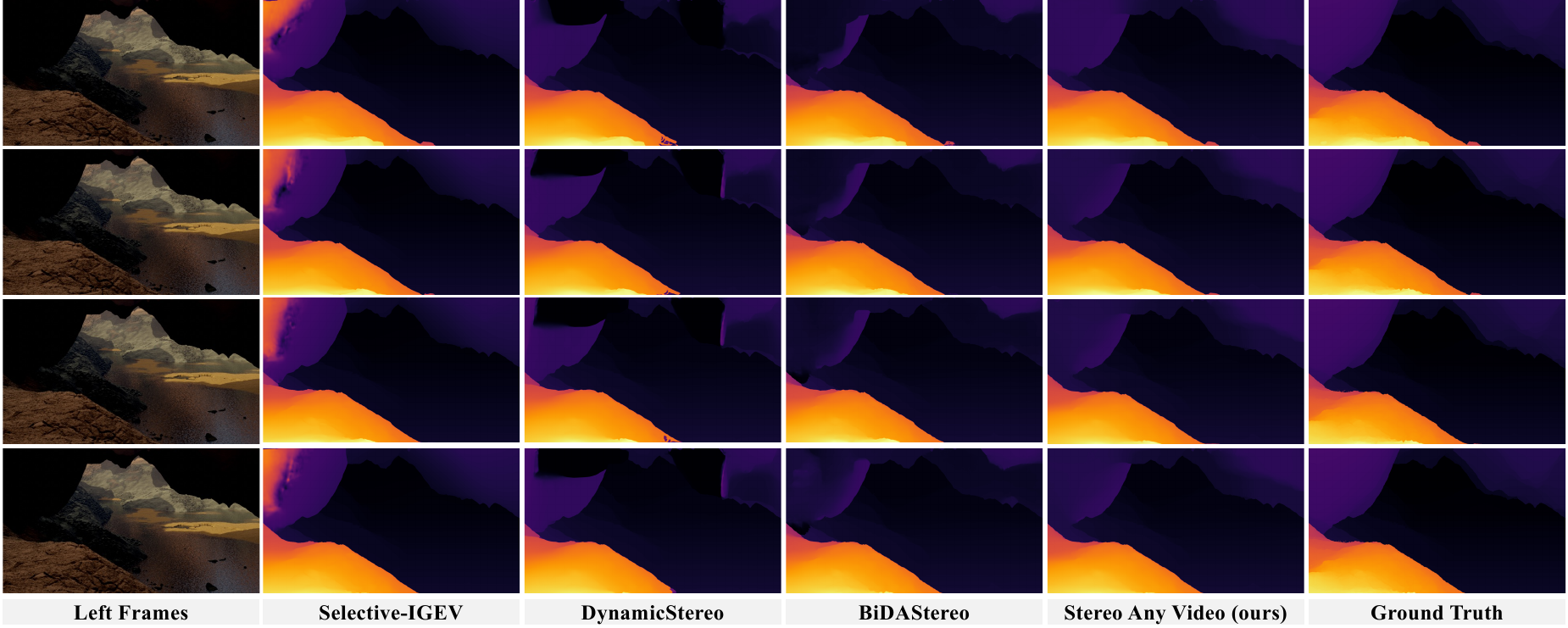}
   \end{center}
   \vspace{-1.2em}
    \caption{Qualitative comparison on Infinigen SV dataset \cite{jing2024matchstereovideosbidirectional}.} 
    \vspace{-.5em}
    \label{fig:sup_1c84cf89}
\end{figure*}

\begin{figure*}[t]
   \begin{center}
   \includegraphics[width=1\linewidth]{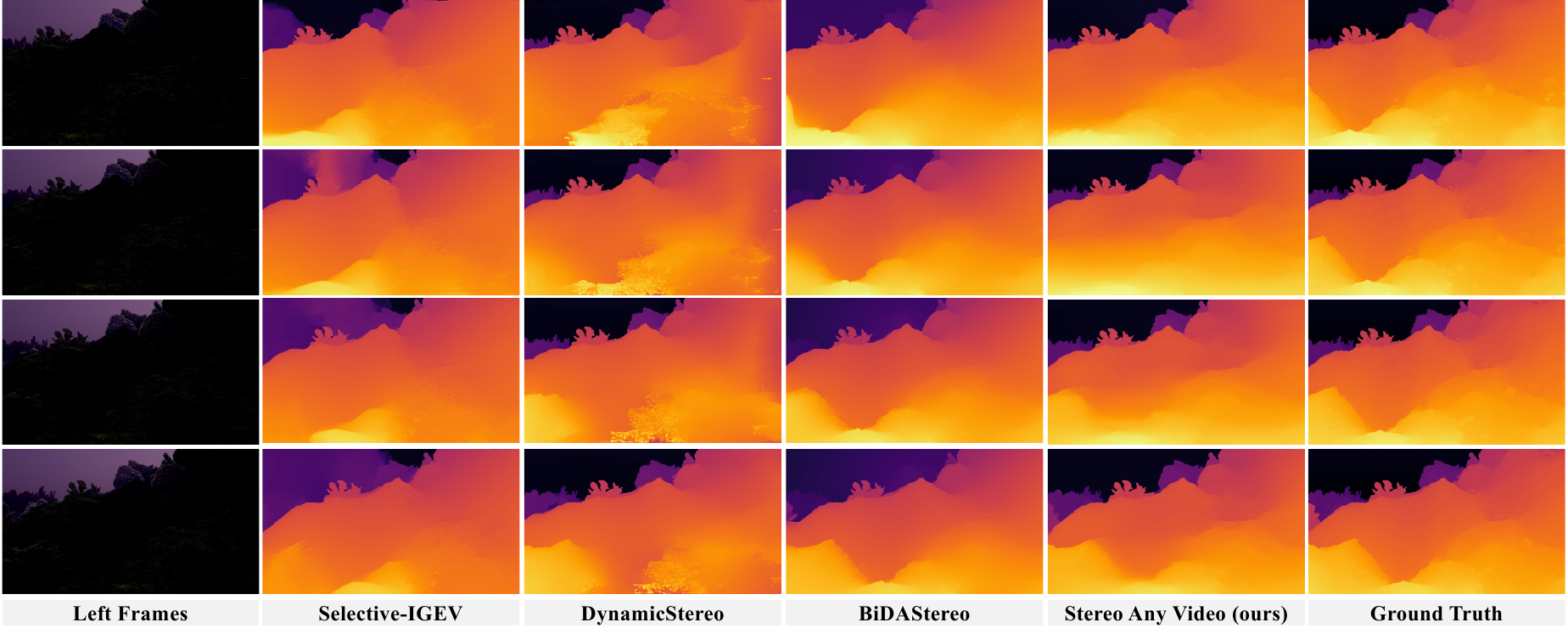}
   \end{center}
   \vspace{-1.2em}
    \caption{Qualitative comparison on Infinigen SV dataset \cite{jing2024matchstereovideosbidirectional}.} 
    \vspace{-.5em}
    \label{fig:sup_23fddf93}
\end{figure*}

\begin{figure*}[t]
   \begin{center}
   \includegraphics[width=1\linewidth]{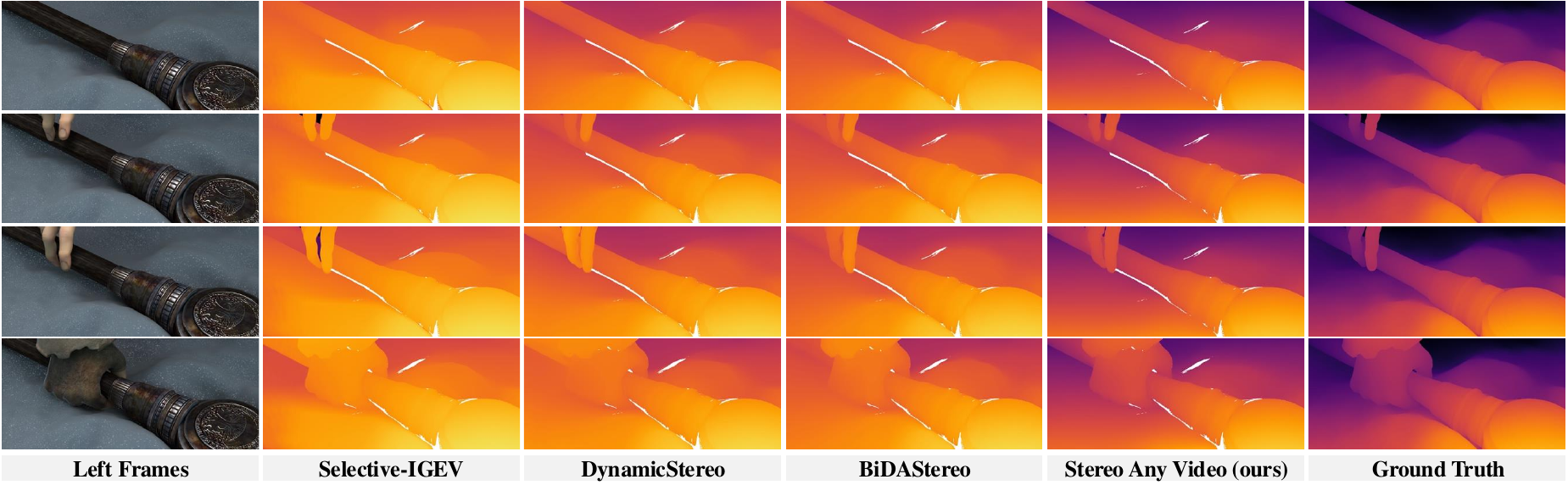}
   \end{center}
   \vspace{-1.2em}
    \caption{Qualitative comparison on Sintel dataset \cite{sintel}.} 
    \vspace{-.5em}
    \label{fig:sup_ambush_7}
\end{figure*}

\begin{figure*}[t]
   \begin{center}
   \includegraphics[width=1\linewidth]{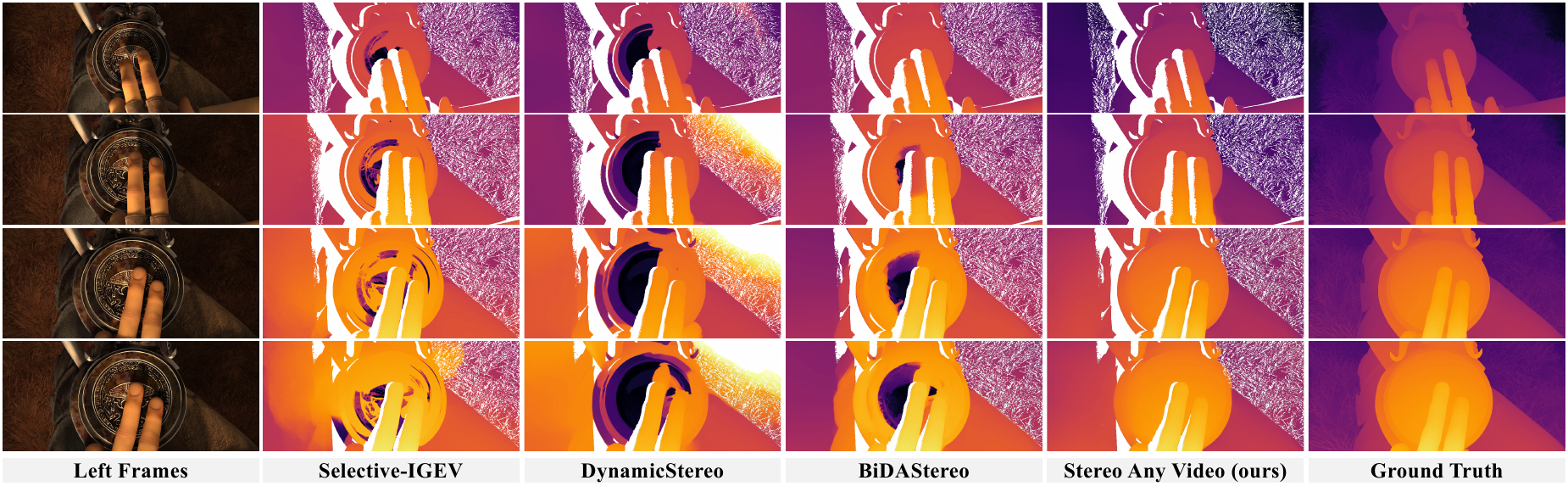}
   \end{center}
   \vspace{-1.2em}
    \caption{Qualitative comparison on Sintel dataset \cite{sintel}.} 
    \vspace{-.5em}
    \label{fig:sup_shaman_3}
\end{figure*}

\begin{figure*}[h]
   \begin{center}
   \includegraphics[width=1\linewidth]{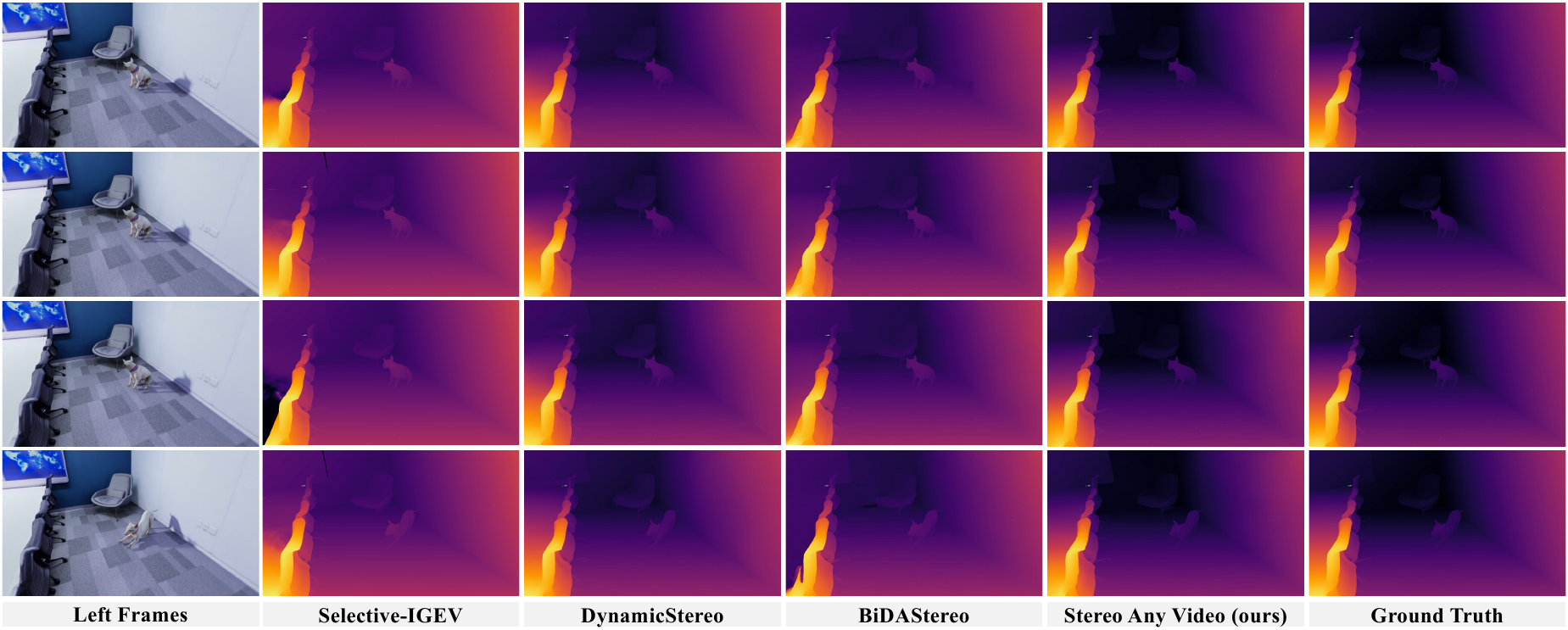}
   \end{center}
   \vspace{-1.2em}
    \caption{Qualitative comparison on Dynamic Replica dataset\cite{karaev2023dynamicstereo}.} 
    \vspace{-.5em}
    \label{fig:sup_01f258-3_obj}
\end{figure*}

\Cref{fig:sup_Scene20} presents the comparison results on Virtual KITTI2 dataset. \Cref{fig:sup_1af03d8d}, \Cref{fig:sup_1c84cf89}, and \Cref{fig:sup_23fddf93} give comparison results on Infinigen SV dataset. \Cref{fig:sup_ambush_7} and  \Cref{fig:sup_shaman_3} show visualization  comparisons on Sintel dataset, and \Cref{fig:sup_01f258-3_obj} shows the results on Dynamic Replica.

\end{document}

%% file: preamble.tex
\usepackage{graphicx}
\usepackage{booktabs}

\usepackage{color}
\usepackage{bm}
\usepackage{multirow}
\usepackage{flushend}
\usepackage{caption}
\usepackage{listings}
\usepackage{caption}